\documentclass[lettersize,journal]{IEEEtran}

\usepackage[table,xcdraw]{xcolor} 
\usepackage{multirow}
\usepackage{booktabs}
\usepackage{arydshln}
\usepackage{array}
\usepackage{pifont}
\definecolor{lightorange}{RGB}{242,242,242}
\definecolor{psnr}{RGB}{255,242,230}
\definecolor{red}{RGB}{204,0,0}
\definecolor{blue}{RGB}{0,102,204}

\usepackage{cuted}
\usepackage{capt-of}
\usepackage{amsmath,amsfonts,amssymb}
\usepackage{bm}
\usepackage{dsfont}
\usepackage{bbding}
\usepackage{cases}
\usepackage{gensymb}
\usepackage{algorithmic}
\usepackage[algoruled,vlined,linesnumbered]{algorithm2e}

\usepackage{graphicx}
\usepackage[caption=false,font=footnotesize]{subfig}

\usepackage{textcomp}
\usepackage{stfloats}
\usepackage{url}
\usepackage{verbatim}
\usepackage{cite}
\usepackage{enumerate}
\usepackage{balance}
\usepackage{comment}
\usepackage{hyperref}

\usepackage{amssymb}
\usepackage{tcolorbox}
\tcbuselibrary{theorems}
\usepackage{lipsum}
\usepackage{enumitem}

\usepackage{placeins}


\usepackage{amsthm}
\theoremstyle{definition} 

\def\BibTeX{{\rm B\kern-.05em{\sc i\kern-.025em b}\kern-.08em
    T\kern-.1667em\lower.7ex\hbox{E}\kern-.125emX}}

\begin{document}
\title{Degradation Frequency Curve: An Explicit Frequency-Quantified Representation for All-in-One Image Restoration}

\author{Xinghua~Huang\textsuperscript{*}, %
        Zhixiong~Yang\textsuperscript{*},~\IEEEmembership{Student Member,~IEEE},
        Chen~Wu,
    Shengxi~Li,~\IEEEmembership{Member,~IEEE},
    Shuaifeng~Zhi,~\IEEEmembership{Member,~IEEE},
        Yue~Zhang,
        Qibin~Hou,~\IEEEmembership{Member,~IEEE},
        Xin~Deng,~\IEEEmembership{Member,~IEEE},
        and Jingyuan~Xia\textsuperscript{\dag},~\IEEEmembership{Member,~IEEE}

\IEEEcompsocitemizethanks{
\IEEEcompsocthanksitem Xinghua~Huang, Zhixiong~Yang, Chen~Wu, Shuaifeng~Zhi, and Jingyuan~Xia are with the College of Electronic Science, National University of Defense Technology, Changsha, 410073, China. \protect\ Shengxi~Li, Yue~Zhang, and Xin~Deng are with the College of Electronic Engineering, Beihang University, Beijing, 100191, China. Qibin~Hou is with VCIP, School of Computer Science, Nankai University, Tianjin, 300071, China.
\IEEEcompsocthanksitem E-mail: (huangxinghua, yzx21, wuchen5, j.xia10)@nudt.edu.cn,\protect\ (LiShengxi, yue\_zhang, cindydeng)@buaa.edu.cn,\protect\ zhishuaifeng@outlook.com,\protect\ houqb@nankai.edu.cn.
\IEEEcompsocthanksitem 
Xinghua~Huang and Zhixiong~Yang contributed equally to this work (Corresponding author: Jingyuan~Xia). \protect
}
\thanks{This work is supported by the National Natural Science Foundation of
China, projects 625B2180, 62576350, 62131020, 62322121, 62376283, and
62531026.}
}

\markboth{Journal of \LaTeX\ Class Files,~Vol.~14, No.~8, August~2021}%
{Shell \MakeLowercase{\textit{et al.}}: A Sample Article Using IEEEtran.cls for IEEE Journals}


\IEEEaftertitletext{%
\begin{center}
  \vspace{-0.4cm}
  \includegraphics[width=\linewidth]{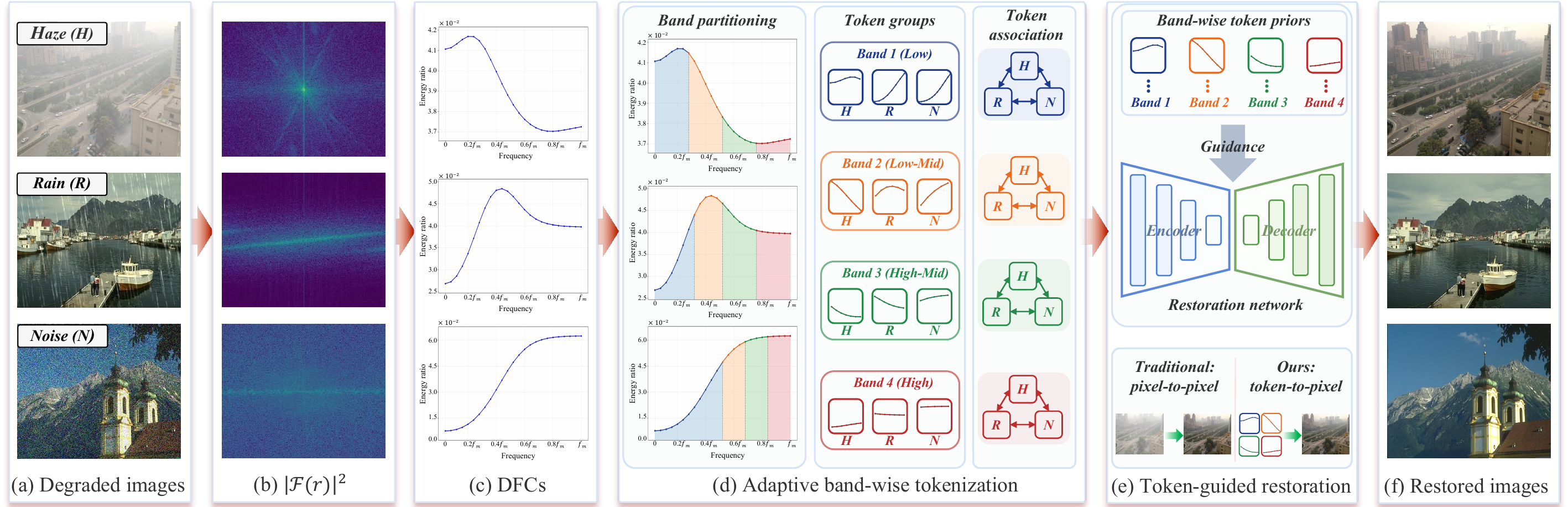}
  \vspace{0.001cm}
  \captionof{figure}{Motivation of the proposed DFC-guided restoration paradigm. 
  (a) Degraded images (from top to bottom: haze, rain, and noise); 
  (b) The corresponding frequency-domain energy maps; 
  (c) The corresponding DFCs obtained by quantifying residual-to-degraded energy ratios;
  (d) Each DFC is adaptively partitioned into several frequency-band segments, forming band-wise spectral tokens. Same-band tokens from different degradations are grouped to learn cross-degradation associations and form compositional token priors;
  (e) The band-wise spectral tokens are fed into the restoration network as degradation-aware priors to guide the restoration process;
  (f) Restored images.}
  \label{fig: DFC_intro}
\end{center}%
}

\maketitle

\begin{abstract}
A fundamental difficulty in all-in-one blind image restoration is that degradation is usually treated as an implicit factor hidden in degraded-to-clean mapping, rather than as an explicit object that can be measured and manipulated. This limitation becomes more pronounced under mixed, compound, or unseen degradation conditions, where degradation effects are hard to assign to predefined labels or task-specific parameters. We propose the Degradation Frequency Curve (DFC), a structured spectral representation that quantifies degradation responses by measuring band-wise residual-to-degraded energy ratios in the frequency domain. DFC converts visually entangled and hard-to-describe degradation effects into a measurable degradation coordinate space. Moreover, DFC can be adaptively decomposed into band-wise spectral tokens, allowing local degradation responses to be represented as reusable restoration priors. Based on this representation, we develop the DFC-guided Image Restorer (DFC-IR), a token-conditioned multi-scale framework that progressively estimates DFCs from intermediate restorations and uses the resulting spectral tokens to guide degradation-aware restoration in a coarse-to-fine manner. Extensive experiments on standard, composite, unseen, and real-world degradation benchmarks show that DFC provides an effective representation basis for all-in-one restoration, leading to state-of-the-art performance and improved generalization under complex degradation profiles.

\end{abstract}


\newtheorem{definition}{Definition}[section]
\newtheorem{theorem}{Theorem}[section]
\newtheorem{assumption}[theorem]{Assumption}
\newtheorem{corollary}{Corollary}[section]
\newtheorem{proposition}{Proposition}[section]
\newtheorem{remark}{Remark}[section]
\newtheorem{lemma}{Lemma}[section]

\section{Introduction}\label{sec:introduction}

\IEEEPARstart{R}{EAL-WORLD} images are frequently degraded by complex and overlapping degradations, such as noise \cite{huang2021neighbor2neighbor, lin2023unsupervised}, blur \cite{cho2021rethinking, kupyn2019deblurgan}, and rain \cite{chen2021robust, jiang2020multi}, posing a fundamental challenge for all-in-one image restoration. Most existing methods, regardless of whether they pursue feature-space unification or parameter specialization, still largely follow a direct degraded-to-restored mapping paradigm. In such a pixel-to-pixel formulation, degradation information is implicitly absorbed into network features rather than being explicitly represented, making the learned restoration process prone to degradation entanglement and inaccurate assignment under complex, compound, or unseen cases. Therefore, a key missing component in all-in-one image restoration is an explicit degradation representation that can transform diverse degradations into a unified, structured, and learnable representation space for degradation-aware restoration.

Current methods typically fall into two categories. Feature-space unification methods \cite{yang2024all,potlapalli2023promptir,conde2024instructir} attempt to learn shared representations for diverse degradations through contrastive or prompt-based learning, but often suffer from feature entanglement and limited degradation discrimination. Parameter-specialization methods \cite{li2022all,park2023all,wang2023smartassign} allocate dedicated subnetworks, experts, or parameter sets for different degradation types, yet they usually introduce additional model complexity and remain vulnerable to inaccurate degradation assignment under ambiguous or compound degradations. Beyond these method-specific limitations, a more fundamental issue remains: both streams still leave degradation modeling largely implicit; the former encodes degradation cues into shared latent features, while the latter ties them to degradation-dependent parameters or branches. Crucially, neither stream explicitly formulates degradation itself as a quantifiable and structured representation, which would allow it to be measured, compared, and used to guide degradation-aware restoration.

Our approach is motivated by a key empirical observation, as shown in Fig. \ref{fig: DFC_intro}: while different degradations are visually entangled and difficult to disentangle in the pixel-level domain (Fig. \ref{fig: DFC_intro}(a)), they exhibit structured and distinguishable patterns in the frequency domain (Fig. \ref{fig: DFC_intro}(b)). However, such spectral differences remain implicit in raw frequency energy maps and are difficult to quantify or directly utilize. To bridge this gap, we introduce the Degradation Frequency Curve (DFC), an explicit, quantifiable, and interpretable spectral representation that characterizes degradation responses across frequency bands, as illustrated in Fig. \ref{fig: DFC_intro}(c). By computing the energy ratios between residual and degraded images across discrete frequency bands, the DFC constructs a spectral degradation coordinate space in which different degradation types exhibit distinguishable response patterns. This representation establishes an explicit, quantifiable, and interpretable basis for degradation awareness, making otherwise elusive degradation patterns observable and measurable in the frequency domain.

Beyond mere characterization, the DFC enables a fundamental shift in how degradations are represented and processed. Although the DFC provides a quantifiable band-wise description of degradation-induced spectral variations, using the entire curve as a single holistic representation overlooks the frequency-dependent nature of degradation responses, where different spectral bands may encode distinct local patterns. Therefore, as illustrated in Fig. \ref{fig: DFC_intro}(d), we introduce adaptive band-wise tokenization, which decomposes each DFC into a set of localized spectral tokens through band partitioning along the frequency axis, as indicated by the color-coded band segments. Each token corresponds to a specific frequency band and jointly encodes its frequency location, local response intensity, and local curve pattern. More importantly, tokens from different degradation types within the same frequency band form band-wise token groups and can be associated with each other, allowing the model to capture both shared spectral primitives and degradation-specific variations. This tokenized representation converts the global DFC into a compositional set of reusable band-wise degradation units, enabling complex or unseen degradations to be modeled through flexible token association and compositional token priors, thereby improving the model's generalization to degradations beyond the training set.

To leverage the tokenized DFC representation for restoration, we develop a DFC-guided Image Restorer (DFC-IR), a multi-scale restoration framework that uses band-wise DFC tokens to guide image restoration, as illustrated in Fig. \ref{fig: DFC_intro}(e). Rather than relying solely on a direct pixel-to-pixel mapping, DFC-IR converts band-wise DFC tokens into restoration priors and performs restoration in a token-to-pixel manner. Specifically, within each scale transition of the coarse-to-fine decoder, a scale-specific DFC is obtained to describe the current degradation response, and the Frequency Curve-guided Mask Sampling (FCMS) module adaptively converts it into band-wise spectral tokens. These tokens are then fed into the Multi-band Modulated Decoding (MMD) module, which learns token-conditioned band-wise representations and adaptively aggregates them for restoration. In this way, DFC-IR translates band-wise DFC tokens into effective restoration guidance within a unified multi-scale framework, without relying on degradation-type labels or handcrafted priors.

The main contributions are summarized as follows:
\begin{itemize}
\item We propose DFC, an explicit, quantifiable, and interpretable spectral representation that characterizes degradation response patterns in the frequency domain and transforms spatially entangled degradations into a spectral degradation coordinate space for degradation-aware restoration.

\item We introduce adaptive DFC tokenization, which decomposes DFCs into band-wise spectral tokens via adaptive frequency partitioning. This tokenized representation preserves band-specific response patterns and supports flexible band-wise token association, thereby providing compositional token priors for complex and unseen degradations.

\item We develop DFC-IR, a multi-scale DFC-guided restoration framework built upon the DFC representation. DFC-IR progressively obtains scale-specific DFCs and translates the resulting band-wise DFC tokens into restoration priors for degradation-aware image restoration.

\item Extensive experiments on standard all-in-one, composite, unseen-degradation, and real-world benchmarks demonstrate that the proposed framework achieves state-of-the-art restoration performance and strong generalization to complex and unseen degradations.
\end{itemize}

\section{Related Work}\label{sec:Related Work}
\subsection{All-in-one Image Restoration}
All-in-one image restoration aims to handle diverse degradation types using a single unified model. Early representative frameworks such as AirNet \cite{li2022all} demonstrated that a shared network can effectively process various degradations without training separate networks for each degradation type. Recent methods \cite{potlapalli2023promptir,conde2024instructir,wang2024gridformer,zamfir2025complexity,cui2025adair,zeng2025vision,tian2025degradation,guo2024onerestore} mainly extend this line in two directions. The first introduces degradation-related conditioning into a shared restoration backbone, such as learnable prompts in PromptIR \cite{potlapalli2023promptir}, human instructions in InstructIR \cite{conde2024instructir}, scene descriptors in OneRestore \cite{guo2024onerestore}, and vision-language guidance in VLU-Net \cite{zeng2025vision}. The second enhances a unified model with adaptive specialization, for example through frequency-aware modulation in AdaIR \cite{cui2025adair}, degradation-aware feature perturbation in DFPIR \cite{tian2025degradation}, and complexity experts in MoCE-IR \cite{zamfir2025complexity}. Together, these designs improve the flexibility of a unified model by either enriching its conditioning signals or adapting its processing pathways to different degradations. Despite their effectiveness, these methods primarily improve all-in-one restoration through conditioning strategies or architectural design, while the explicit characterization of degradation remains less explored.

\subsection{Degradation Representation}
A key challenge in all-in-one image restoration is characterizing diverse degradations to guide the network. Existing methods mainly address this issue by introducing degradation-related cues to support restoration. Early methods extract degradation priors directly from inputs, with AirNet \cite{li2022all} learning unified latent representations and NDR \cite{yao2024neural} encoding continuous neural representations for feature modulation. Subsequent studies incorporate prompt-based or embedding-based guidance: PromptRestorer \cite{wang2023promptrestorer} and DA-CLIP \cite{luo2023controlling} extract degradation-aware embeddings, while MPerceiver \cite{ai2024multimodal} and UniProcessor \cite{duan2024uniprocessor} employ multimodal or task-specific prompts. Recently, DCPT \cite{hu2025universal} leverages degradation-type classification for pre-training. Overall, existing methods predominantly encode degradation as implicit guidance variables, leaving the degradation itself abstract and insufficiently characterized. In contrast, our DFC explicitly formulates degradation as a compact spectral representation, providing a frequency-motivated and interpretable basis for guiding the restoration process.

\begin{figure*}[t!]
\vspace{0cm}
  \centering
  \includegraphics[width=1\linewidth]{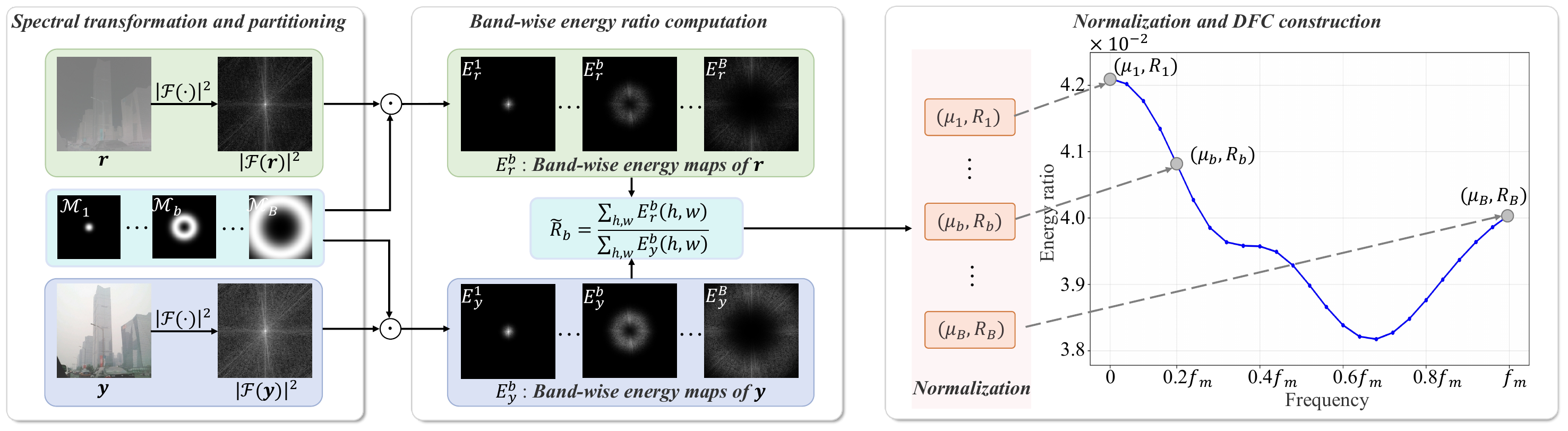}\\
  \vspace{0cm}
    \caption{Overview of DFC construction.
The DFC is obtained by transforming the residual and degraded images into the frequency domain, computing band-wise energy ratios with Gaussian masks, and normalizing the ratios to form the final curve.}
  \label{DFC_generate}
  \vspace{0cm}
\end{figure*}

\subsection{Frequency-aware Image Restoration}
Frequency-domain processing has emerged as a prominent paradigm in image restoration. Early studies primarily focused on fixed spectral transformations, such as reconstructing missing high-frequency details via local spectral estimation \cite{yoo2018image}, conducting representation learning directly within the Fourier space \cite{xu2020learning}, and optimizing spectral objectives like Focal Frequency Loss \cite{jiang2021focal}. Recent advancements have shifted toward adaptive frequency manipulation. For instance, selective frequency mechanisms \cite{cui2023selective,cui2023image} have been proposed to dynamically decompose and route informative spectral bands, while AdaIR \cite{cui2025adair} further introduces frequency-aware modulation for all-in-one restoration. These studies collectively demonstrate the effectiveness of frequency-domain cues for image restoration, ranging from spectral reconstruction and spectral optimization to adaptive feature modulation. However, these methods mainly exploit frequency information at the feature level, without explicitly isolating degradation characteristics in the spectral domain, which fundamentally limits their restoration performance under complex degradations. In contrast, our DFC explicitly characterizes degradation in the frequency domain as a compact band-wise representation for restoration.

\section{Degradation Frequency Curve Modeling}
\label{sec:dfc_modeling}
The core of our approach is the DFC, a frequency-domain representation that characterizes image degradation through band-wise spectral responses. By representing degradation as an explicit band-wise spectral curve, DFC provides an interpretable and compact prior for all-in-one image restoration.

\subsection{DFC Formulation}
The DFC is defined as a $B$-dimensional vector $D(f) \in \mathbb{R}^{B}$ extracted from a degraded image $\bm{y} \in \mathbb{R}^{H \times W \times 3}$ and its clean counterpart $\bm{x} \in \mathbb{R}^{H \times W \times 3}$, where $B$ denotes the number of frequency bands and $f$ denotes the frequency. Each element in $D(f)$ represents the normalized degradation response in a specific frequency band, and the whole curve provides a compact spectral description of the degradation. To extract this spectral description, we first compute the residual image $\bm{r} = \bm{y} - \bm{x}$ in the pixel space to isolate the degradation component. As shown in Fig.~\ref{DFC_generate}, the DFC computation takes $\bm{r}$ and $\bm{y}$ as inputs and consists of the following three steps. 

\textbf{Step 1: Spectral transformation and partitioning.}
We perform two preparatory operations in the frequency domain:
1) obtaining the spectral energy maps of both $\bm{r}$ and $\bm{y}$ using the operator $|\mathcal{F}(\cdot)|^2$, where $\mathcal{F}(\cdot)$ denotes the Fourier transform; and
2) partitioning the frequency domain into $B$ contiguous bands using a set of predefined ring-shaped Gaussian masks $\{\mathcal{M}_b\}_{b=1}^B$.
The $b$-th mask is defined as:
\begin{equation}
\mathcal{M}_b(h,w)=\exp\left(-\frac{(d(h,w)-\mu_b)^2}{2\sigma_b^2}\right),
\label{eq: pre-defined Gaussian masks}
\end{equation}
where $d(h,w)$ denotes the $\ell_2$ distance from a pixel $(h,w)$ to the DC component in the frequency domain, $\mu_b$ is the central frequency of the $b$-th band, and $\sigma_b$ controls its bandwidth. Unlike rigid frequency partitioning, Gaussian masks introduce smooth transitions between adjacent bands via soft assignment, which reduces sensitivity to band boundaries and enhances the robustness of spectral decomposition against continuous spectral variations caused by diverse degradations. To ensure comprehensive spectral coverage, $\mu_b$ increases linearly from $0$ to the maximum radial frequency $\frac{\sqrt{H^2+W^2}}{2}$, while $\sigma_b$ increases linearly from $0.05$ to $0.3$. This progressively increasing bandwidth design is motivated by the non-uniform spectral statistics of degraded images: in the energy-dominant low-frequency regions where degradation-related responses are more pronounced, narrower bands facilitate fine-grained discrimination of spectral responses. Conversely, in the sparser high-frequency regions, wider bands help maintain the stability of energy estimation and prevent over-fragmented partitioning that may be susceptible to stochastic noise.

\begin{figure*}[t!]
\vspace{-0.15cm}
  \centering
  \includegraphics[width=1\linewidth]{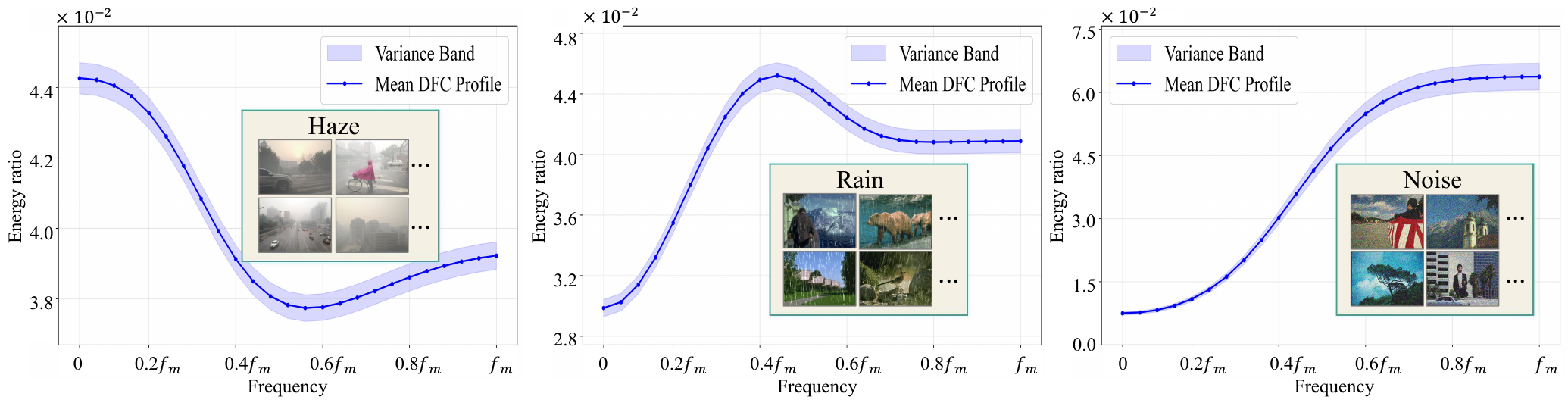}\\
  \vspace{0cm}
  \caption{Statistical DFC profiles for haze, rain, and noise. The solid lines and colored bands represent the mean profiles and variance bands, respectively.}
  \label{fig:statistical_dfc}
  \vspace{-0.15cm}
\end{figure*}

\textbf{Step 2: Band-wise energy ratio computation.}
To quantify the degradation response within each frequency band, we first extract the band-wise energy components by applying the predefined Gaussian masks to the spectral energy maps of $\bm{r}$ and $\bm{y}$:
\begin{equation}
E_r^{b}=\mathcal{M}_b \odot |\mathcal{F}(\bm{r})|^2,
\end{equation}
\begin{equation}
E_y^{b}=\mathcal{M}_b \odot |\mathcal{F}(\bm{y})|^2,
\end{equation}
where $\odot$ is element-wise multiplication. Subsequently, the energy ratio $\tilde{R}_b$ for the $b$-th frequency band is calculated as:
\begin{equation}
\tilde{R}_b=\frac{\sum_{h,w} E_r^{b}(h,w)}{\sum_{h,w} E_y^{b}(h,w)}.
\end{equation}
The adoption of energy ratios rather than absolute energy values essentially treats $E_y^b$ as the observed spectral energy baseline within the same frequency band for relative quantification. Such a design helps reduce the influence of image content, thereby improving the robustness of the degradation representation across diverse scenes.

\textbf{Step 3: Normalization and DFC construction.}
Finally, to obtain the DFC, the energy ratio set $\{\tilde{R}_b\}_{b=1}^B$ is normalized to a unit sum:
\begin{equation}
R_b=\frac{\tilde{R}_b}{\sum_{j=1}^{B}\tilde{R}_j}.
\end{equation}
This step transforms the discrete band responses into a probability-like distribution along the frequency axis, shifting the focus of the representation toward the relative allocation of degradation energy across bands and its spectral distribution patterns. The normalized results are then mapped to their corresponding central frequencies $\{\mu_b\}_{b=1}^B$ to yield the $B$-dimensional vector $D(f)$. Through these sequential operations, the spatially entangled degradation is explicitly converted into a compact and discriminative 1D spectral curve.

\subsection{Properties of DFC}
$D(f)$ exhibits three key properties for degradation representation: content robustness, degradation sensitivity, and severity-aware response. These properties show that DFC can provide a stable and discriminative description of degradation-related frequency responses.

\textbf{Content robustness.}
A desirable degradation representation should be less dominated by image content, since the same degradation may appear on diverse scenes with substantially different structures and textures. To verify this property, we randomly sample 100 images for each degradation type and visualize their mean DFC profiles together with the corresponding variance bands. As shown in Fig.~\ref{fig:statistical_dfc}, the DFCs of the same degradation remain within a relatively narrow variance range despite variations in image content. This indicates that the energy-ratio formulation suppresses part of content-dependent spectral variations and highlights degradation-related band-wise frequency responses. Therefore, DFC provides a content-robust representation for the same degradation across different visual contents.

\textbf{Degradation sensitivity.}
Complementary to content robustness, degradation sensitivity reflects the ability of DFC to distinguish different degradation types through their spectral response patterns. Fig.~\ref{fig:statistical_dfc} shows that different degradations exhibit clearly distinct mean DFC profiles. Specifically, haze mainly induces responses in low-frequency regions, rain produces more prominent responses in middle-frequency regions, while noise is concentrated in high-frequency regions. Such discriminability is not limited to the location of the dominant frequency band, but is also reflected in the peak position, response concentration, and overall curve shape. These observations confirm the degradation sensitivity of DFC, showing that different degradation types form distinguishable and interpretable spectral response patterns in the DFC space.

\begin{figure*}[t!]
    \centering

    \subfloat[Noise DFC profiles]{
        \includegraphics[height=2.6cm]{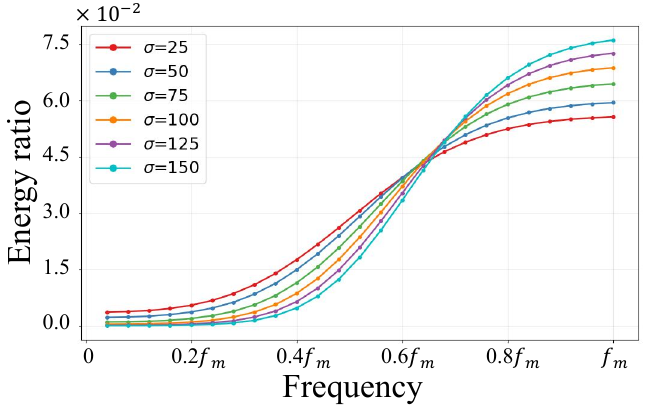}
        \label{fig:dfc_severity_a}
    }
    \hfil
    \subfloat[Noise peak response]{
        \includegraphics[height=2.6cm]{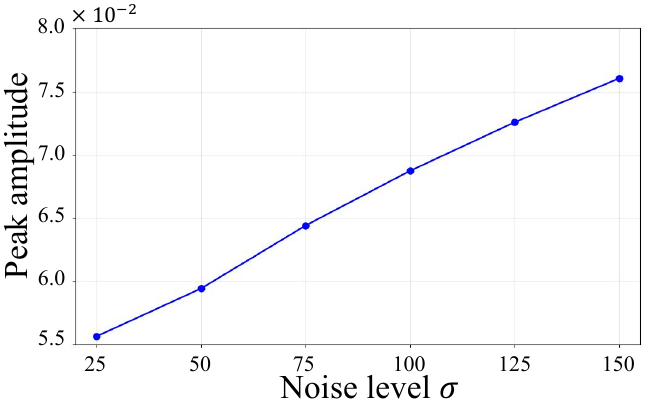}
        \label{fig:dfc_severity_b}
    }
    \hfil
    \subfloat[Low-light DFC profiles]{
        \includegraphics[height=2.6cm]{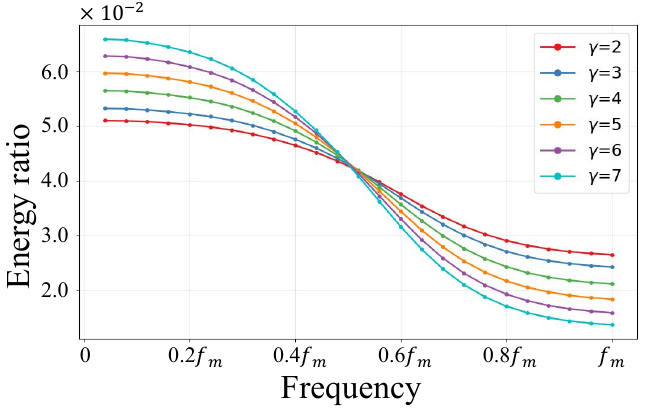}
        \label{fig:dfc_severity_c}
    }
    \hfil
    \subfloat[Low-light peak response]{
        \includegraphics[height=2.6cm]{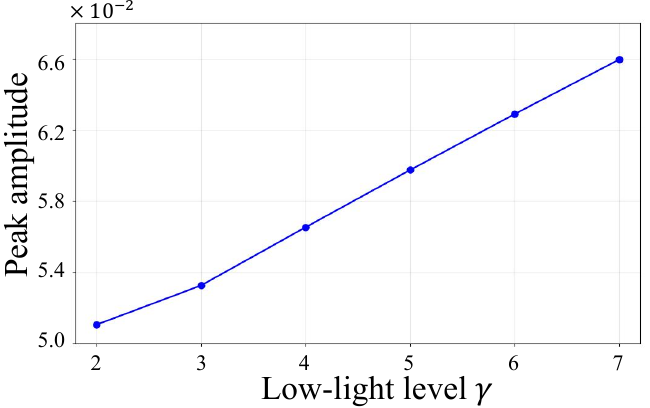}
        \label{fig:dfc_severity_d}
    }

    \vspace{0.2cm}
    \caption{Severity-aware DFC variation. 
    (a)(b) DFC profiles and peak response variations under different Gaussian noise levels $\sigma$. 
    (c)(d) DFC profiles and peak response variations under different low-light levels $\gamma$.}
    \label{fig:dfc_severity}
\end{figure*}

\begin{figure*}[t!]
    \centering

    \subfloat[Band-wise tokenization of DFCs]{
        \includegraphics[height=5.66cm]{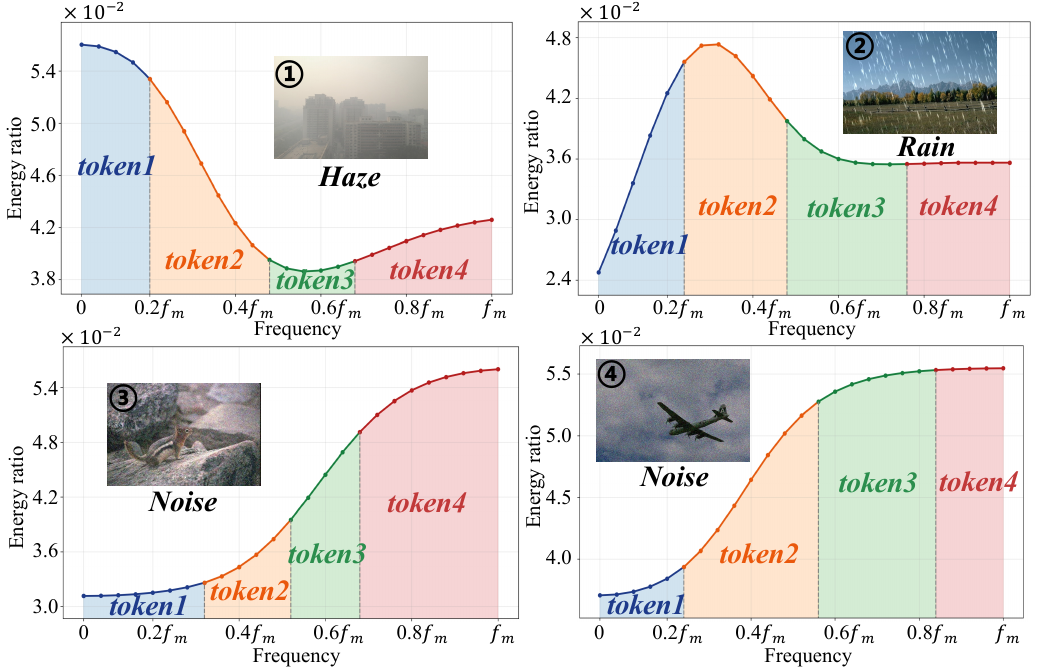}
        \label{fig:dfc_tokenization_a}
    }
    \hfil
    \subfloat[Compositional token priors for unseen degradations]{
        \includegraphics[height=5.66cm]{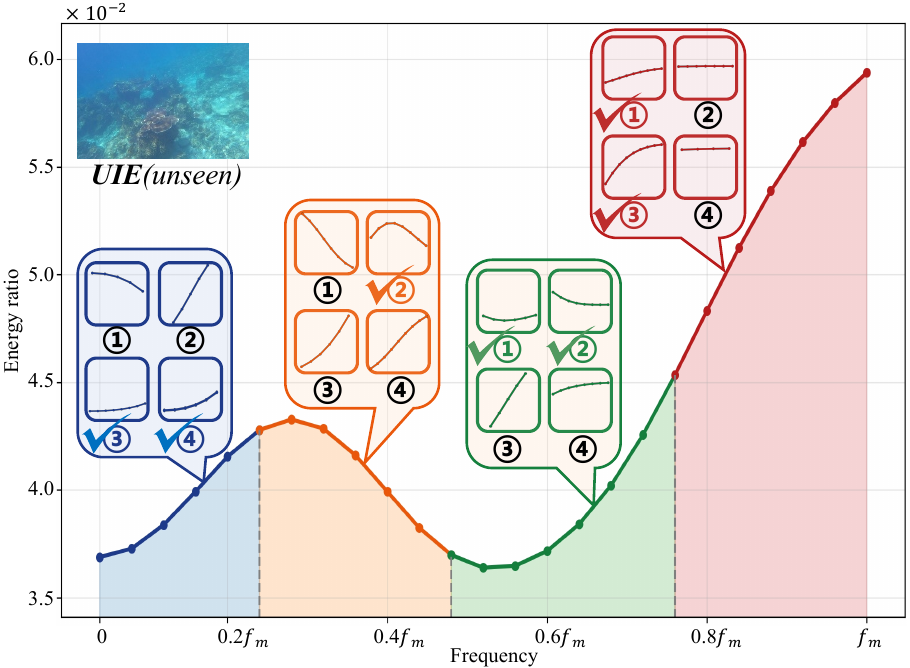}
        \label{fig:dfc_tokenization_b}
    }

    \vspace{0.2cm}
    \caption{Illustration of DFC tokenization and compositional token priors for unseen degradations. 
    (a) Seen degradation DFCs are decomposed into band-wise spectral tokens. 
    (b) The DFC of an unseen underwater degradation can be characterized through reusable band-wise token priors from seen DFCs. 
    The marked same-band tokens share similar local response patterns with the corresponding frequency regions of the unseen DFC and jointly approximate its local spectral profile, illustrating the compositional spectral basis provided by tokenized DFCs.}
    \label{fig:dfc_tokenization}
\end{figure*}

\textbf{Severity-aware response.}
In addition to distinguishing degradation types, DFC is also sensitive to degradation severity. As shown in Fig.~\ref{fig:dfc_severity}, we analyze Gaussian noise with different noise levels $\sigma$ and low-light degradation with different illumination parameters $\gamma$. In both cases, increasing the degradation level leads to a systematic change in the DFC response. The overall curve shape remains relatively consistent within the same degradation family, while the peak amplitude changes monotonically with degradation severity. This indicates that DFC not only characterizes degradation-type differences, but also captures continuous severity-related variations within each degradation type.

\subsection{Decomposing DFC into Spectral Tokens}
Based on the above properties, we use $D(f)$ as an explicit degradation prior to provide the restoration model with frequency-aware cues. However, the informative cues encoded in DFC are inherently frequency-dependent. Directly using the entire $D(f)$ as a single global condition is too coarse, since it compresses the responses of different frequency bands into one holistic representation. This will obscure local spectral variations and weaken band-specific degradation cues, resulting in less targeted restoration modulation. To preserve these local responses, we introduce DFC tokenization, which decomposes the global curve into a sequence of band-wise spectral tokens.

As illustrated in Fig.~\ref{fig:dfc_tokenization}(a), the DFCs of different degraded samples, including haze, rain, and noise cases, are partitioned into multiple tokens along the frequency axis. Each token corresponds to a local frequency segment and encodes the response intensity and local curve pattern within that segment. In this way, DFC tokenization converts the global degradation curve into a set of localized band-wise representations, making the frequency-dependent structure of degradation more explicit. The detailed tokenization strategy will be introduced in Sec.~\ref{sec:4.2}.

Beyond preserving localized frequency responses, DFC tokenization also provides compositional token priors for unseen degradations. Although an unseen degradation can exhibit a new global DFC profile, its local frequency responses can share common patterns with those of seen degradations. As illustrated in Fig.~\ref{fig:dfc_tokenization}(b), the DFC of an unseen underwater degradation can be characterized through band-wise token priors obtained from the tokenized DFCs of the seen degradation samples in Fig.~\ref{fig:dfc_tokenization}(a). For example, the third token of the UIE DFC shares similar local response patterns with the marked third tokens of the haze and rain DFCs, and these same-band tokens can jointly approximate the corresponding frequency region of the unseen DFC. This compositional token-prior view indicates that tokenized DFCs provide a reusable band-wise spectral basis, where local degradation patterns from seen degradations collectively contribute to characterizing complex or unseen degradations, thereby enabling more flexible degradation modeling beyond the seen degradation types.

\section{DFC-Guided Image Restoration}\label{sec:method}
\begin{figure*}[t!]
  \vspace{0cm}
  \centering
  \includegraphics[width=1\linewidth]{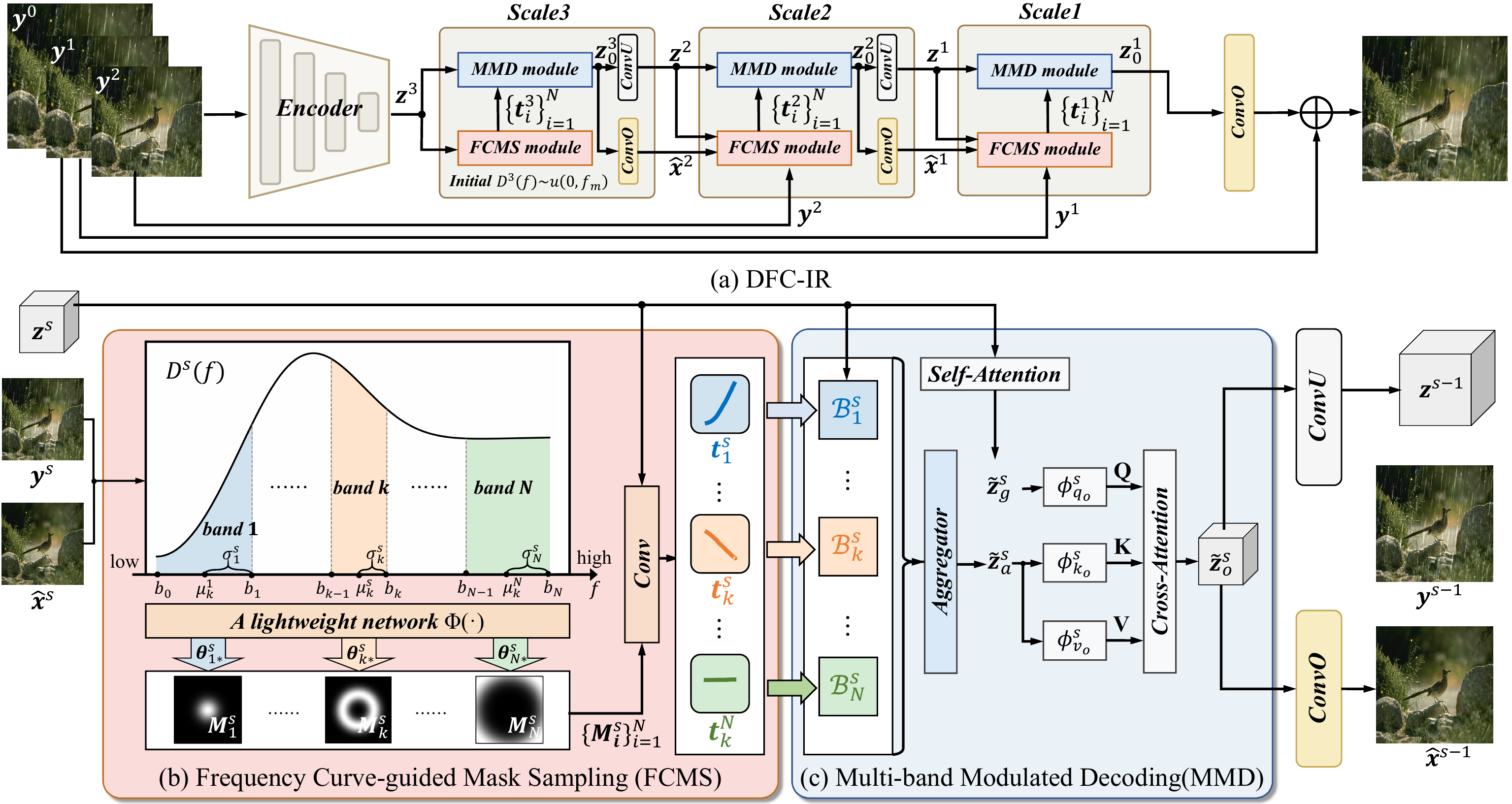}\\
  \vspace{0.2cm}
\caption{The architecture of the proposed DFC-IR. 
(a) The overall multi-scale encoder-decoder framework. 
(b) Adaptive DFC tokenization implemented by the FCMS module, which adaptively partitions the scale-specific DFC and extracts band-wise spectral tokens.
(c) Band-wise token decoding implemented by the MMD module, which uses DFC tokens as band-wise priors for feature modulation and adaptive aggregation.}
\label{DPP-Net}
  \vspace{0cm}
\end{figure*}

\subsection{Overview of DFC-IR}
\label{sec:4.1}
DFC-IR leverages DFCs as explicit spectral priors for all-in-one image restoration. As shown in Fig.~\ref{DPP-Net}(a), DFC-IR adopts a multi-scale encoder-decoder architecture that takes a three-level degraded image pyramid $\bm{Y}=\{\bm{y}^i\}_{i=0}^{2}$ as input. The pyramid is constructed by downsampling the original degraded image $\bm{y}^0$ by factors of 2 and 4, where the $i$-th scale has spatial dimensions $\mathbb{R}^{\frac{H}{2^i}\times\frac{W}{2^i}\times 3}$. The encoder extracts a latent feature $\bm{z}^3$ from $\bm{Y}$, which encodes both image content and degradation information across scales.

The core of DFC-IR lies in its DFC-guided decoding stage, which progressively refines the restoration from low to high resolution through three scale loops $(s=3,2,1)$. At each scale transition, the decoder performs a DFC-token-conditioned restoration process. Specifically, a scale-specific DFC \(D^s(f)\) is first obtained to describe the current degradation response. For the coarsest scale, since a reliable intermediate restoration is not yet available, \(D^s(f)\) is initialized as a uniform curve. For the following scales, \(D^s(f)\) is estimated from the current intermediate restoration $\hat{\bm{x}}^s$ and the degraded input $\bm{y}^s$. Then, the Frequency Curve-guided Mask Sampling (FCMS) module implements adaptive DFC tokenization by transforming $D^s(f)$ into a set of band-wise spectral tokens extracted from the current latent feature $\bm{z}^s$:
\begin{equation}
\{\bm{t}_i^s\}_{i=1}^{N}=
\mathcal{T}_{\mathrm{FCMS}}\left(\bm{z}^s,D^s(f)\right),
\label{eq:fcms_tokenization}
\end{equation}
where \(\mathcal{T}_{\mathrm{FCMS}}\) denotes the adaptive DFC tokenization operator implemented by FCMS.

Subsequently, these DFC tokens are fed into the Multi-band Modulated Decoding (MMD) module for DFC-token-conditioned feature modulation and restoration:
\begin{equation}
\bm{z}^{s-1}, \hat{\bm{x}}^{s-1}
=
\mathcal{R}_{\mathrm{MMD}}\left(\bm{z}^s,\{\bm{t}_i^s\}_{i=1}^{N}\right),
\label{eq:mmd_restoration}
\end{equation}
where \(\mathcal{R}_{\mathrm{MMD}}\) denotes the DFC-token-conditioned decoding operator implemented by MMD. The resulting feature \(\bm{z}^{s-1}\) and intermediate restoration \(\hat{\bm{x}}^{s-1}\) are passed to the next finer scale, where the DFC is re-estimated and tokenized again. In this way, DFC-IR forms a coarse-to-fine restoration loop that progressively estimates scale-specific DFCs, decomposes them into band-wise spectral tokens, and converts these tokens into band-wise priors for feature modulation and degradation-aware restoration.

\subsection{Adaptive DFC Tokenization}
\label{sec:4.2}
At each scale, DFC-IR decomposes the scale-specific DFC \(D^s(f)\) into a set of band-wise spectral tokens for subsequent band-wise token decoding. As shown in Fig.~\ref{DPP-Net}(b), this adaptive tokenization process is implemented by FCMS, which determines frequency partitions according to \(D^s(f)\) and extracts the corresponding spectral tokens from the current latent feature \(\bm{z}^s\). Specifically, FCMS partitions the frequency domain using a set of Gaussian masks $\{\bm{M}_i^s\}_{i=1}^N$ defined in Eq.~\ref{eq: pre-defined Gaussian masks}. To balance the degradation responses carried by different tokens, we divide \(D^s(f)\) into \(N\) equal-area regions and estimate the corresponding mask parameters $\{\hat{\bm{\theta}}_i^s=(\hat{\mu}_i^s,\hat{\sigma}_i^s)\}_{i=1}^{N}$. Fundamentally, this equal-area division balances the degradation response mass rather than the geometric frequency span. This strategy prevents highly imbalanced token informativeness, ensuring that each generated token encapsulates a relatively informative spectral segment rather than an overly dominant or nearly empty band. For the $i$-th region, $\hat{\mu}_i^s$ and $\hat{\sigma}_i^s$ are initialized as its central frequency and bandwidth, respectively.

However, this initial partition is still limited in adapting to complex and diverse degradations. Therefore, FCMS further uses a parameter sampling mechanism to refine these initial estimates, thereby enabling adaptive frequency partitioning. Specifically, for the $i$-th mask, we sample $K$ parameters $\{\bm{\theta}_{i,k}^s=(\mu_{i,k}^s,\sigma_{i,k}^s)\}_{k=1}^{K}$ around $\hat{\bm{\theta}}_i^s$. A lightweight network $\Phi(\cdot)$ is then used to predict a weight distribution over these $K$ sampled parameters:
\begin{equation}
    \{\omega_{i,k}^s\}_{k=1}^K = \text{softmax}(\Phi\left(\{\bm\theta_{i,k}^s\}_{k=1}^K\right)).
\end{equation}
The refined Gaussian parameters $\bm{\theta}_{i*}^s=(\mu_{i*}^s,\sigma_{i*}^s)$ are aggregated via weighted summation:
\begin{equation}
\mu_{i*}^s = \sum_{k=1}^K \omega_{i,k}^s \mu_{i,k}^s, \quad \sigma_{i*}^s = \sum_{k=1}^K \omega_{i,k}^s \sigma_{i,k}^s.
\end{equation}
Through this weighted aggregation, the frequency partition becomes more flexible around the initial equal-area partition, allowing the masks to better capture localized degradation responses.

Based on the refined parameters $\{\bm{\theta}_{i*}^s\}_{i=1}^{N}$, we construct the Gaussian masks $\{\bm{M}_i^s\}_{i=1}^{N}$ using Eq.~\ref{eq: pre-defined Gaussian masks}. Finally, to explicitly encode the band-specific degradation priors, we apply these masks to the latent feature $\bm{z}^s$ in the frequency domain and extract the spectral tokens $\{\bm{t}_i^s\}_{i=1}^{N}$ via a set of learnable convolution blocks. The token $\bm{t}_i^s$ for the $i$-th band is formulated as follows:
\begin{equation}
\bm{t}_i^s = \text{conv}_i\left(\mathcal{F}^{-1}(\mathcal{F}(\text{conv}_s(\bm{z}^s)) \odot \bm{M}_i^s)\right),
\label{frequency_prototypes}
\end{equation}
where $\text{conv}_s$ is a shared convolution, $\text{conv}_i$ is the band-specific convolution for the $i$-th frequency band, $\mathcal{F}$ and $\mathcal{F}^{-1}$  denote the Fourier and inverse Fourier transforms, and $\odot$ represents element-wise multiplication. Semantically, each \(\bm{t}_i^s\) serves as a DFC-guided band-wise spectral token, which couples the localized spectral response indicated by \(D^s(f)\) with the current latent feature \(\bm{z}^s\), thereby providing band-specific cues for subsequent restoration modulation.

\subsection{Band-wise Token Decoding}
\label{sec:4.3}
At each scale, the DFC tokens \(\{\bm{t}_i^s\}_{i=1}^{N}\) are used as band-wise priors to condition the current latent feature \(\bm{z}^s\), enabling degradation cues from different frequency bands to guide the decoding process. As shown in Fig.~\ref{DPP-Net}(c), this process is implemented by MMD, which performs token-conditioned feature modulation and adaptive aggregation. To exploit the band-wise cues encoded in DFC tokens, MMD employs a set of band-wise modulation branches \(\{\mathcal{B}_i^s\}_{i=1}^{N}\). Since degradation responses exhibit distinct characteristics across the frequency spectrum, a single shared transformation may dilute band-specific restoration requirements. In contrast, each modulation branch is conditioned on its corresponding token \(\bm{t}_i^s\), enabling the input feature \(\bm{z}^s\) to be refined according to the degradation response of a specific frequency band. Specifically, the \(i\)-th branch uses \(\bm{t}_i^s\) as a DFC token prior to modulate \(\bm{z}^s\) through cross-attention:
\begin{equation}
\tilde{\bm{z}}_i^s =
\mathrm{softmax}
\left(
\frac{
\phi_{q_i}^s(\bm{z}^s)
\left(\phi_{k_i}^s(\bm{t}_i^s)\right)^{T}
}{\sqrt{d}}
\right)
\phi_{v_i}^s(\bm{t}_i^s),
\end{equation}
where \(\phi_{q_i}^s\), \(\phi_{k_i}^s\), and \(\phi_{v_i}^s\) are linear projections. Through this cross-attention mechanism, each DFC token serves as a band-wise degradation prior for its corresponding modulation branch, producing a token-conditioned representation \(\tilde{\bm{z}}_i^s\).

The token-conditioned modulation outputs are then fused by a lightweight adaptive aggregator \(\mathcal{G}(\cdot)\), which predicts the aggregation weights from the DFC tokens:
\begin{equation}
\{\alpha_i^s\}_{i=1}^{N}
=
\mathrm{softmax}
\left(
\mathcal{G}\left(\{\bm{t}_i^s\}_{i=1}^{N}\right)
\right).
\end{equation}
These weights are used to aggregate the token-conditioned modulation outputs:
\begin{equation}
\tilde{\bm{z}}_a^s
=
\sum_{i=1}^{N}
\alpha_i^s \tilde{\bm{z}}_i^s .
\end{equation}
Since the DFC tokens encode sample-specific frequency responses, this token-aware aggregation allows the decoder to emphasize the most relevant band-wise representations for the current degradation.

Although DFC tokens enable band-wise feature modulation, independently processing different token-conditioned features may weaken cross-band coherence. To preserve global structural consistency, we derive a global context feature \(\tilde{\bm{z}}_g^s\) by applying self-attention to the input feature \(\bm{z}^s\). This global feature then interacts with the aggregated band-wise feature through cross-attention:
\begin{equation}
\bm{z}_o^s =
\mathrm{softmax}
\left(
\frac{
\phi_{q_o}^s(\tilde{\bm{z}}_g^s)
\left(\phi_{k_o}^s(\tilde{\bm{z}}_a^s)\right)^{T}
}{\sqrt{d}}
\right)
\phi_{v_o}^s(\tilde{\bm{z}}_a^s).
\end{equation}
This interaction harmonizes the locally refined band-wise representations with global structural context, maintaining visual coherence in the restored output. The enhanced feature \(\bm{z}_o^s\) is upsampled to \(\bm{z}^{s-1}\) via a convolutional upsampling block (ConvU), and processed by a convolutional output block (ConvO) to produce \(\hat{\bm{x}}^{s-1}\), thereby completing the decoding step from scale \(s\) to \(s-1\).

\subsection{Training Objective}
\label{Sec:4.4}

DFC-IR is optimized in an end-to-end manner with multi-scale supervision. Following the DFC-guided restoration process described in Sec.~\ref{sec:4.1}--\ref{sec:4.3}, each scale-specific DFC \(D^s(f)\) is tokenized by FCMS and then used to condition the MMD decoder. Accordingly, we formulate the objective directly on the restoration outputs guided by \(D^s(f)\), so that the learned DFC tokens are optimized as band-wise restoration priors.

To preserve spatial-domain restoration accuracy, we apply a spatial reconstruction loss to the DFC-guided outputs at all scales:
\begin{equation}
\mathcal{L}_{s}
=
\sum_{s\in\{1,2,3\}}
\left\|
\mathcal{R}^{x}_{\mathrm{MMD}}
\left(
\bm{z}^{s},
\mathcal{T}_{\mathrm{FCMS}}
\left(
\bm{z}^{s},D^{s}(f)
\right)
\right)
-
\bm{x}^{s-1}
\right\|_{1},
\end{equation}
where \(\mathcal{R}^{x}_{\mathrm{MMD}}\) denotes the image-output branch of the MMD decoder. This loss encourages the DFC-guided output at each scale to recover faithful spatial structures and details with respect to the clean target.

Furthermore, we introduce a frequency-domain consistency loss to reduce spectral discrepancies between the DFC-guided restoration and the corresponding clean target:
\begin{equation}
\begin{aligned}
\mathcal{L}_{f}
=
\sum_{s\in\{1,2,3\}}
\Big\|
&\mathcal{F}\Big(
\mathcal{R}^{x}_{\mathrm{MMD}}
\big(
\bm{z}^{s},
\mathcal{T}_{\mathrm{FCMS}}(\bm{z}^{s},D^{s}(f))
\big)
\Big)
\\
&-
\mathcal{F}
\big(
\bm{x}^{s-1}
\big)
\Big\|_{1},
\end{aligned}
\end{equation}
where \(\mathcal{F}(\cdot)\) denotes the Fourier transform. This term complements the spatial supervision by enforcing spectral consistency between the restored output and the clean target, further aligning the DFC-guided restoration process with the frequency-domain structure of the clean target.

The total objective is formulated as:
\begin{equation}
\mathcal{L}
=
\lambda_s \mathcal{L}_{s}
+
\lambda_f \mathcal{L}_{f},
\end{equation}
where \(\lambda_s\) and \(\lambda_f\) are weighting hyperparameters. The scale-specific DFCs used in these losses are computed in the same way during training and inference, as described in Sec.~\ref{sec:4.1}.
\section{Experiments}
Following previous state-of-the-art works \cite{li2022all,potlapalli2023promptir,zhang2025perceive, zhang2023ingredient}, we conduct extensive experiments to validate our DFC-IR. The evaluations are primarily performed under two standard settings: (i) the all-in-one setting, where a unified model is trained to address multiple degradations; and (ii) the composite degradation setting, where a single model is trained to restore images suffering from a complex combination of multiple degradations simultaneously. Furthermore, to comprehensively assess the generalization capability of our method, we extend our analysis to unseen degradations and complex real-world scenarios. We evaluate pixel-level fidelity using PSNR and SSIM \cite{wang2004image}, measure perceptual quality via LPIPS \cite{zhang2018unreasonable} and FID \cite{heusel2017gans}, and assess real-world images with three no-reference metrics: MANIQA \cite{yang2022maniqa}, CLIPIQA \cite{wang2023exploring}, and MUSIQ \cite{ke2021musiq}. The best and second-best results are \textbf{highlighted} and \underline{underlined}, respectively.

\subsection{Experimental Setup}
\subsubsection{Dataset Preparation} 
For the all-in-one setting, our experimental setup aligns with established practices \cite{li2022all, potlapalli2023promptir}. Specifically, for image denoising, we integrate the BSD400 \cite{arbelaez2010contour} and WED \cite{ma2016waterloo} datasets, adding Gaussian noise at levels $\sigma\in\{15,25,50\}$ to generate corrupted images, and test on the BSD68 \cite{martin2001database}. For deraining and dehazing, we adopt the Rain100L \cite{yang2017deep} and SOTS \cite{li2018benchmarking} datasets, respectively. Deblurring and low-light enhancement experiments are conducted on the GoPro \cite{nah2017deep} and LOL-v1 \cite{wei2018deep} datasets. To construct the unified all-in-one model, we merge the training data from the aforementioned datasets to form two configurations: one with three degradations and another with five. For the composite degradation setting, we conduct experiments on the CDD11 \cite{guo2024onerestore} dataset to evaluate the model's capability in effectively handling complex mixed degradations.

\subsubsection{Implementation Details} 
DFC-IR is implemented as an end-to-end trainable 4-level encoder-decoder framework. The encoder comprises four levels, progressively stacked with [4, 6, 6, 8] transformer blocks \cite{zamir2022restormer,potlapalli2023promptir} from the top to the bottom. The decoder includes three levels, configured with [2, 4, 4] instances of each FCMS and MMD module across the respective stages. The importance estimator $\Phi(\cdot)$ in FCMS is implemented as a lightweight three-layer MLP, with the number of sampled parameters $K$ empirically set to 5. The number of predefined masks, $B$, for DFC construction is set to 25. The number of band-wise DFC tokens $N$ in the FCMS module is set to 4, and the MMD module uses the corresponding number of band-wise modulation branches. For training, we trained our models for 100 epochs with batch size 32 using Adam \cite{kingma2014adam} ($\beta_1=0.9$, $\beta_2=0.999$) at initial learning rate $2\times 10^{-4}$ with cosine decay. During training, we employ $128 \times 128$ randomly cropped patches and augment the data with random horizontal and vertical flipping, while optimizing a combined $\mathcal{L}_1$ loss in both RGB and Fourier domains with hyperparameters $\lambda_s=1$ and $\lambda_f=0.1$. All experiments are conducted on 4 NVIDIA A800 GPUs.

\subsection{Comparison to State-of-the-Art Methods}
Following the experimental settings of prior works \cite{potlapalli2023promptir, conde2024instructir, zamfir2025complexity}, we compare our DFC-IR with state-of-the-art methods, including general image restoration models such as DGUNet \cite{mou2022deep}, MPRNet \cite{zamir2021multi}, Restormer \cite{zamir2022restormer}, and MambaIR \cite{guo2024mambair}, as well as specialized all-in-one methods like AirNet \cite{li2022all}, WGWSNet \cite{zhu2023learning}, WeatherDiff \cite{ozdenizci2023restoring}, PromptIR \cite{potlapalli2023promptir}, Gridformer \cite{wang2024gridformer}, InstructIR \cite{conde2024instructir}, OneRestore \cite{guo2024onerestore}, Art-PromptIR \cite{wu2024harmony}, DA-RCOT \cite{tang2025degradation}, AdaIR \cite{cui2025adair}, VLU-Net \cite{zeng2025vision}, MoCE-IR \cite{zamfir2025complexity}, and DFPIR \cite{tian2025degradation}.

\textbf{Three degradations}. The initial evaluation is conducted across three degradation types: haze, rain, and Gaussian noise at noise levels $\sigma \in \{15, 25, 50\}$. Quantitative results in Table~\ref{tab:3task} demonstrate that the proposed DFC-IR consistently achieves superior performance compared to existing approaches. Specifically, DFC-IR obtains an average PSNR gain of 0.12 dB over the state-of-the-art method DFPIR \cite{tian2025degradation} and a 0.27 dB improvement over MoCE-IR \cite{zamfir2025complexity}.

\begin{table*}[ht]
\centering
\small
\setlength{\tabcolsep}{3.5pt} 
\caption{Quantitative comparison (PSNR/SSIM) of all-in-one image restoration on the three-degradation setting. The best and second-best results are \textbf{highlighted} and \underline{underlined}, respectively.}
\vspace{0cm}
\resizebox{\linewidth}{!}{\begin{tabular}{@{} ccc c cccccccc cccc @{}} 
\toprule
\multirow{2}{*}{{Type}} & 
\multirow{2}{*}{{Method}} & 
\multirow{2}{*}{{Reference}} & 
\multirow{2}{*}{{Params.}} & 
\multicolumn{2}{c}{{Dehazing}} & 
\multicolumn{2}{c}{{Deraining}} & 
\multicolumn{6}{c}{{Denoising}} & 
\multicolumn{2}{c}{{Average}} \\
\cmidrule(lr){5-6} \cmidrule(lr){7-8} \cmidrule(lr){9-14} \cmidrule(lr){15-16}
 & & & & \multicolumn{2}{c}{{SOTS}} & \multicolumn{2}{c}{{Rain100L}} & \multicolumn{2}{c}{{BSD68}$_{\sigma=15}$} & \multicolumn{2}{c}{{BSD68}$_{\sigma=25}$} & \multicolumn{2}{c}{{BSD68}$_{\sigma=50}$} & PSNR & SSIM \\
\midrule

\multirow{3}{*}{General} 
& MPRNet \cite{zamir2021multi} & CVPR'21 & 16.3M & 28.00 & .958 & 33.86 & .958 & 33.27 & .920 & 30.76 & .871 & 27.29 & .761 & 30.63 & .894 \\
& Restormer \cite{zamir2022restormer} & CVPR'22 & 26.1M & 27.78 & .958 & 33.78 & .958 & 33.72 & .930 & 30.67 & .865 & 27.63 & .792 & 30.75 & .901 \\
& MambaIR \cite{guo2024mambair} & ECCV'24 & 27.2M & 29.57 & .970 & 35.42 & .969 & 33.88 & .931 & 30.95 & .874 & 27.74 & .793 & 31.51 & .907 \\

\midrule

\multirow{8}{*}{All-in-one}
& Gridformer \cite{wang2024gridformer} & IJCV'24 & 34.1M & 30.37 & .970 & 37.15 & .972 & 33.93 & .931 & 31.37 & .887 & 28.11 & {.801} & 32.19 & .912 \\

& Art-PromptIR \cite{wu2024harmony} & ACMMM'24 & 33.2M & 30.83 & .979 & 37.94 & .982 & 34.06 & \underline{.934} & 31.42 & .891 & 28.14 & .801 & 32.49 & .917 \\

& InstructIR \cite{conde2024instructir} & ECCV'24 & 15.8M & 30.22 & .959 & {37.98} & {.978} & \underline{34.15} & .933 & \underline{31.52} & {.890} & \textbf{28.30} & \underline{.804} & {32.43} & {.913} \\

& DA-RCOT \cite{tang2025degradation} & TPAMI'25 & 52.3M & 31.26 & .977 & {38.36} & {.983} & {33.98} & \underline{.934} & {{31.33}} & {.890} & {{28.10}} & {.801} & {32.60} & {.917} \\

& AdaIR \cite{cui2025adair} & ICLR'25 & 28.8M & 31.06 & \underline{.980} & 38.64 & .983 & 34.12 & \textbf{.935} & 31.45 & \underline{.892} & 28.19 & .802 & 32.69 & \underline{.918} \\

& VLU-Net \cite{zeng2025vision} & CVPR'25 & 35.4M & {30.71} & \underline{.980} & \textbf{38.93} & \underline{.984} & 34.13 & \textbf{.935} & 31.48 & \underline{.892} & 28.23 & \underline{.804} & {32.70} & \textbf{.919} \\

& MoCE-IR \cite{zamfir2025complexity} & CVPR'25 & 25.4M & 31.34 & .979 & 38.57 & \underline{.984} & 34.11 & .932 & 31.45 & .888 & 28.18 & .800 & 32.73 & .917 \\

& DFPIR \cite{tian2025degradation} & CVPR'25 & 31.1M & \underline{31.87} & \underline{.980} & 38.65 & .982 & 34.14 & \textbf{.935} & 31.47 & \textbf{.893} & 28.25 & \textbf{.806} & \underline{32.88} & \textbf{.919} \\

\cmidrule(lr){2-16} 
& {DFC-IR} & 
{Ours} & 
33.1M & 
\textbf{32.28} & 
\textbf{.982} & 
\underline{38.74} & 
\textbf{.985} & 
\textbf{34.19} & 
{.933} & 
\textbf{31.53} & 
{.890} & 
\underline{28.28} & 
\underline{.804} &
\textbf{33.00} & 
\textbf{.919} \\
\bottomrule
\end{tabular}}
\label{tab:3task}
\end{table*}

\textbf{Five degradations}. Based on \cite{li2022all, zhang2023ingredient}, we extend the three original degradation types to include deblurring and low-light image enhancement, further validating the effectiveness of our method. As shown in Table~\ref{tab:5task}, the proposed DFC-IR achieves a 0.39 dB gain compared to the recent state-of-the-art method DFPIR \cite{tian2025degradation}, when averaged across five restoration tasks. Compared to another competitive method, MoCE-IR \cite{zamfir2025complexity}, the improvement reaches 0.45 dB. 

\begin{table*}[ht]
\centering
\small
\setlength{\tabcolsep}{4.0pt} 
\caption{Quantitative comparison (PSNR / SSIM) for all-in-one restoration on five tasks. }
\resizebox{\linewidth}{!}{\begin{tabular}{@{} ccc c cccccccc cccc @{}} 
\toprule
\multirow{2}{*}{{Type}} & 
\multirow{2}{*}{{Method}} & 
\multirow{2}{*}{{Reference}} & 
\multirow{2}{*}{{Params.}} & 
\multicolumn{2}{c}{{Dehazing}} & 
\multicolumn{2}{c}{{Deraining}} & 
\multicolumn{2}{c}{{Denoising}} & 
\multicolumn{2}{c}{{Deblurring}} & 
\multicolumn{2}{c}{{Low-light}} & 
\multicolumn{2}{c}{{Average}} \\
\cmidrule(lr){5-6} \cmidrule(lr){7-8} \cmidrule(lr){9-10} \cmidrule(lr){11-12} \cmidrule(lr){13-14} \cmidrule(lr){15-16}
 & & & & \multicolumn{2}{c}{{SOTS}} & \multicolumn{2}{c}{{Rain100L}} & \multicolumn{2}{c}{{BSD68}$_{\sigma=25}$} & \multicolumn{2}{c}{{GoPro}} & \multicolumn{2}{c}{{LOLv1}} & PSNR & SSIM \\
\midrule
\multirow{3}{*}{General} 
& DGUNet \cite{mou2022deep}  & CVPR'22 & 17.3M & 24.78 & .940 & 36.62 & .971 & 31.10 & .883 & 27.25 & .837 & 21.87 & .823 & 28.32 & .891 \\
& Restormer \cite{zamir2022restormer} & CVPR'22 & 26.1M & 24.09 & .927 & 34.81 & .962 & \underline{31.49} & .884 & 27.22 & .829 & 20.41 & .806 & 27.60 & .881 \\
& MambaIR \cite{guo2024mambair}  & ECCV'24 & 27.2M & 25.81 & .944 & 36.55 & .971 & 31.41 & .884 & 28.61 & .875 & 22.49 & .832 & 28.97 & .901 \\

\midrule

\multirow{8}{*}{All-in-one}
& PromptIR \cite{potlapalli2023promptir} & NeurIPS'23 & 35.6M & 26.54 & .949 & 36.37 & .970 & {31.47} & .886 & 28.71 & .881 & 22.68 & .832 & 29.15 & .904 \\
& Gridformer \cite{wang2024gridformer}  & IJCV'24 & 34.1M & 26.79 & .951 & 36.61 & .971 & 31.45 & .885 & 29.22 & .884 & 22.59 & .831 & 29.33 & .904 \\
& InstructIR \cite{conde2024instructir}  & ECCV'24 & 15.8M & 27.10 & .956 & 36.84 & .973 & 31.40 & .887 & 29.40 & .886 & 23.00 & .836 & 29.55 & .907 \\
& VLU-Net \cite{zeng2025vision} & CVPR'25 & 35.4M & 30.84 & \underline{.980} & \textbf{38.54} & \underline{.982} & 31.43 & \textbf{.891} & 27.46 & .840 & {22.29} & .833 & 30.11 & .905 \\
& AdaIR \cite{cui2025adair}  & ICLR'25 & 28.8M & 30.53 & .978 & 38.02 & .981 & 31.35 & \underline{.889} & 28.12 & .858 & 23.00 & .845 & 30.20 & .910 \\
& DA-RCOT \cite{tang2025degradation} & TPAMI'25 & 52.3M & 30.96 & .975 & 37.87 & .980 & 31.23 & .888 & 28.68 & .872 & \underline{23.25} & .836 & 30.40 & .911 \\
& MoCE-IR \cite{zamfir2025complexity}  & CVPR'25 & 25.4M & 30.48 & .974 & \underline{38.04} & \underline{.982} & 31.34 & .887 & \textbf{30.05} & \textbf{.899} & 23.00 & \underline{.852} & 30.58 & \underline{.919} \\
& DFPIR \cite{tian2025degradation} & CVPR'25 & 31.1M & \underline{31.64} & .979 & 37.62 & .978 & 31.29 & \underline{.889} & 28.82 & .873 & \textbf{23.82} & .843 & \underline{30.64} & .913 \\
\cmidrule(lr){2-16}
& DFC-IR & Ours & 33.1M & \textbf{32.27} & \textbf{.982} & 37.91 & \textbf{.983} & \textbf{31.55} & \textbf{.891} & \underline{29.62} & \underline{.892} & \textbf{23.82} & \textbf{.863} & \textbf{31.03} & \textbf{.922} \\
\bottomrule
\end{tabular}}
\label{tab:5task}
\end{table*}

\begin{table*}[ht]
\centering
\small
\setlength{\tabcolsep}{2pt} 
\caption{Quantitative evaluation of all-in-one image restoration performance on the CDD11 dataset under complex compounded degradations.}
\vspace{0cm}
\resizebox{\linewidth}{!}{
\begin{tabular}{@{} c c *{12}{cc} @{}} 
\toprule
\multirow{2}{*}{Method} & \multirow{2}{*}{Params.} & 
\multicolumn{8}{c}{{CDD11-Single}} & 
\multicolumn{10}{c}{{CDD11-Double}} & 
\multicolumn{4}{c}{{CDD11-Triple}} & 
\multicolumn{2}{c}{{Average}} \\
\cmidrule(lr){3-10} \cmidrule(lr){11-20} \cmidrule(lr){21-24}
\cmidrule(lr){25-26} && 
\multicolumn{2}{c}{Low (L)} & \multicolumn{2}{c}{Haze (H)} & \multicolumn{2}{c}{Rain (R)} & \multicolumn{2}{c}{Snow (S)} & 
\multicolumn{2}{c}{L+H} & \multicolumn{2}{c}{L+R} & \multicolumn{2}{c}{L+S} & \multicolumn{2}{c}{H+R} & \multicolumn{2}{c}{H+S} & 
\multicolumn{2}{c}{L+H+R} & \multicolumn{2}{c}{L+H+S} & PSNR & SSIM \\
\midrule

AirNet \cite{li2022all} & 9M & 
24.83 & .778 & 24.21 & .951 & 26.55 & .891 & 26.79 & .919 & 
23.23 & .779 & 22.82 & .710 & 23.29 & .723 & 22.21 & .868 & 23.29 & .901 & 
21.80 & .708 & 22.24 & .725 & 
23.75 & .814 \\

PromptIR \cite{potlapalli2023promptir} & 36M & 
26.32 & .805 & 26.10 & .969 & 31.56 & .946 & 31.53 & .960 & 
24.49 & .789 & 25.05 & .771 & 24.51 & .761 & 24.54 & .924 & 23.70 & .925 & 
23.74 & .752 & 23.33 & .747 & 
25.90 & .850 \\

WGWSNet \cite{zhu2023learning} & 26M & 
24.39 & .774 & 27.90 & .982 & 33.15 & .964 & 34.43 & .973 & 
24.27 & .800 & 25.06 & .772 & 24.60 & .765 & 27.23 & .955 & 27.65 & .960 & 
23.90 & .772 & 23.97 & .771 & 
26.96 & .863 \\

WeatherDiff \cite{ozdenizci2023restoring} & 83M & 
23.58 & .763 & 21.99 & .904 & 24.85 & .885 & 24.80 & .888 & 
21.83 & .756 & 22.69 & .730 & 22.12 & .707 & 21.25 & .868 & 21.99 & .868 & 
21.23 & .716 & 21.04 & .698 & 
22.49 & .799 \\

OneRestore \cite{guo2024onerestore} & 6M & 
26.48 & \underline{.826} & 32.52 & \underline{.990} & 33.40 & .964 & 34.31 & .973 & 
25.79 & \underline{.822} & 25.58 & .799 & 25.19 & .789 & \underline{29.99} & .957 & \underline{30.21} & .964 & 
24.78 & .788 & 24.90 & \underline{.791} & 
28.47 & .878 \\

MoCE-IR-S \cite{zamfir2025complexity} & 11M & 
\underline{27.26} & .824 & \underline{32.66} & \underline{.990} & \underline{34.31} & \underline{.970} & \underline{35.91} & \underline{.980} & 
\underline{26.24} & .817 & \underline{26.25} & \underline{.800} & \underline{26.04} & \underline{.793} & 29.93 & \underline{.964} & 30.19 & \underline{.970} & 
\underline{25.41} & \underline{.789} & \underline{25.39} & .790 & 
\underline{29.05} & \underline{.881} \\

\midrule
DFC-IR  & 33.1M & 
\textbf{27.71} & \textbf{.831} & \textbf{34.94} & \textbf{.993} & \textbf{35.41} & \textbf{.976} & \textbf{37.22} & \textbf{.983} & 
\textbf{26.87} & \textbf{.828} & \textbf{26.87} & \textbf{.812} & \textbf{26.85} & \textbf{.806} & \textbf{31.94} & \textbf{.973} & \textbf{32.15} & \textbf{.977} & 
\textbf{26.27} & \textbf{.804} & \textbf{26.25} & \textbf{.803} & 
\textbf{30.22} & \textbf{.890} \\

\bottomrule
\end{tabular}
}
\label{mix}
\end{table*}

\textbf{Composite degradations}. To further validate the effectiveness of the proposed method in addressing composite degradations, experiments are conducted on the CDD11 dataset, which encompasses both single and mixed degradation types. Quantitative results in Table~\ref{mix} demonstrate that DFC-IR consistently achieves superior performance. Specifically, the model obtains a substantial PSNR gain of 1.17 dB compared to the current leading approach MoCE-IR-S \cite{zamfir2025complexity} and yields a notable improvement of 1.75 dB over the recent method OneRestore \cite{guo2024onerestore}. This significant enhancement is primarily attributed to the DFC-guided spectral tokenization strategy, which effectively characterizes mixed spectral degradation responses across distinct frequency bands, thereby enabling token-conditioned band-specific feature modulation and adaptive aggregation for complex mixed degradations.

\begin{figure*}[t!]
  \vspace{0cm}
  \centering
  \includegraphics[width=1\linewidth]{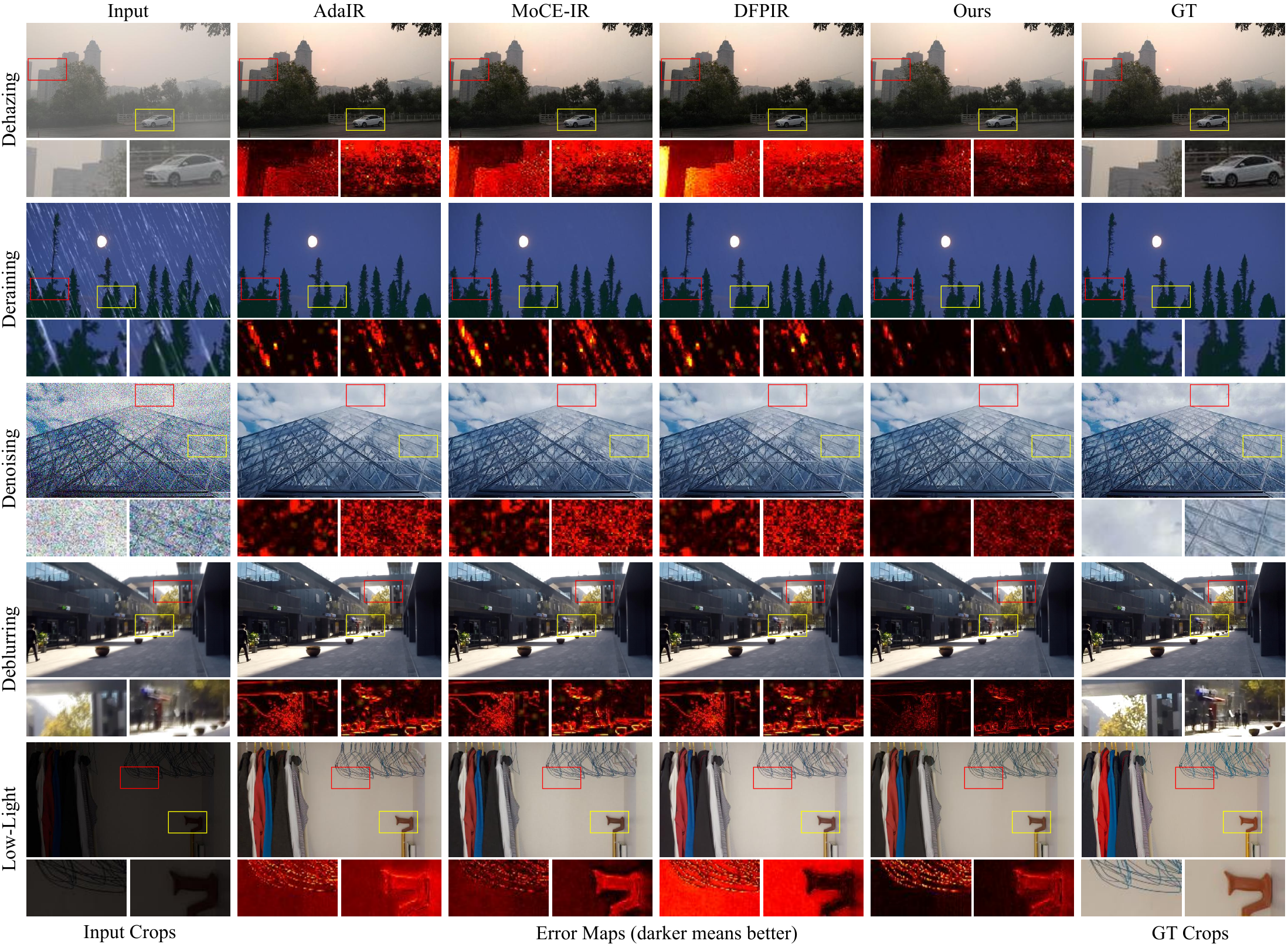}\\
  \vspace{0cm}
  \caption{Visual results for all-in-one restoration on five tasks. The pixel-wise error is visualized using a heatmap, where a color gradient from black to white denotes increasing error magnitude. 
  }\label{fig: visual results}
  \vspace{0cm}
\end{figure*}

\textbf{Visual results.} Fig.~\ref{fig: visual results} presents visual comparisons against recent state-of-the-art methods, including AdaIR \cite{cui2025adair}, MoCE-IR \cite{zamfir2025complexity}, and DFPIR \cite{tian2025degradation}, across five image restoration tasks: dehazing, deraining, denoising, deblurring, and low-light enhancement. In complex hazy scenes, our method achieves more accurate color reconstruction, significantly mitigating the color bias present in existing approaches. For image deraining, our method reliably removes rain streaks and produces visually cleaner results. In denoising tasks, our approach recovers finer structural details while generating sharp, noise-free outputs. For image deblurring, our method restores sharp edges and clear textures from severely blurred inputs, avoiding the ringing artifacts commonly produced by other models. In low-light enhancement, our approach recovers details in dark regions and exhibits superior noise suppression compared to the evaluated baselines. We further provide error maps to demonstrate that our method leaves minimal residual artifacts across all degradations, offering solid evidence of its effectiveness.

\subsection{Generalization to Unseen Degradations}
To comprehensively evaluate the generalization ability of the proposed method, we evaluate its performance on out-of-distribution degradation types (underwater image enhancement and desnowing) as well as unseen degradation levels (unseen rain and noise levels). The quantitative evaluation employs four metrics: PSNR, SSIM \cite{wang2004image}, LPIPS \cite{zhang2018unreasonable}, and FID \cite{heusel2017gans}.

\begin{table}[!t]
\centering
\caption{Quantitative comparison on unseen restoration tasks: UIE on UIEB and desnowing on CSD.}
\label{tab:unseen_types}
\setlength{\tabcolsep}{2.5pt} 
\resizebox{\linewidth}{!}{
\begin{tabular}{@{} l cccc cccc @{}}
\toprule
\multirow{2}{*}{Method} & 
\multicolumn{4}{c}{UIE (UIEB)} & 
\multicolumn{4}{c}{Desnowing (CSD)}  \\
\cmidrule(lr){2-5} \cmidrule(lr){6-9} 
 & PSNR$\uparrow$ & SSIM$\uparrow$ & LPIPS$\downarrow$ & FID$\downarrow$ & PSNR$\uparrow$ & SSIM$\uparrow$ & LPIPS$\downarrow$ & FID$\downarrow$ \\
\midrule
Restormer \cite{zamir2022restormer} & 17.28 & 0.768 & 0.304 & 49.56 & 18.89 & 0.746 & 0.359 & 130.1 \\
PromptIR \cite{potlapalli2023promptir}  & 17.51 & 0.774 & 0.289 & 45.77 & 19.12 & 0.750 & 0.356 & 126.2 \\
DFPIR \cite{tian2025degradation} & 17.74 & 0.771 & 0.296 & 45.01 & 19.78 & 0.759 & 0.337 & 114.7 \\
DA-RCOT \cite{tang2025degradation} & 18.27 & 0.794 & 0.256 & 33.47 & 20.21 & 0.774 & 0.324 & 106.3 \\
AdaIR \cite{cui2025adair} & 18.44 & 0.800 & 0.251 & 32.76 & 20.44 & 0.783 & \underline{0.322} & 101.6 \\
VLU-Net \cite{zeng2025vision} & 18.51 & 0.802 & \underline{0.248} & \underline{30.01} & 20.52 & 0.771 & 0.328 & 94.87 \\
MoCE-IR \cite{zamfir2025complexity} & \underline{18.65} & \underline{0.804} & 0.256 & 33.74 & \underline{20.60} & \underline{0.784} & 0.324 & \underline{92.16} \\
\midrule
DFC-IR (Ours) & \textbf{19.24} & \textbf{0.817} & \textbf{0.231} & \textbf{27.76} & \textbf{21.05} & \textbf{0.802} & \textbf{0.291} & \textbf{70.68} \\
\bottomrule
\end{tabular}
}
\end{table}

\begin{figure*}[t!]
  \vspace{0cm}
  \centering
  \includegraphics[width=1\linewidth]{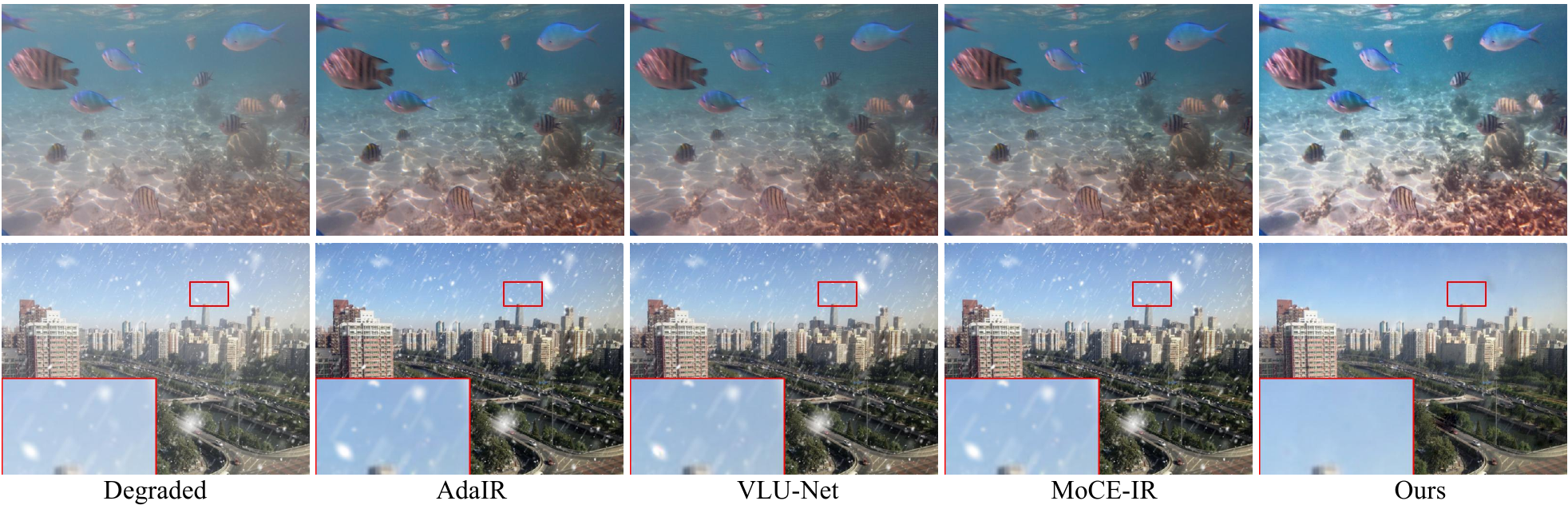}\\
  \vspace{0cm}
  \caption{Visual comparisons on unseen restoration tasks, with UIE on UIEB shown in the first row and desnowing on CSD shown in the second row.
Our method effectively restores accurate underwater colors and removes dense snow particles. 
  }
  \label{fig_unseen_type}
  \vspace{0cm}
\end{figure*}

\textbf{Unseen degradation types.}
We first directly apply the model trained on five seen degradations to two out-of-distribution restoration tasks: underwater image enhancement (UIE) and desnowing. The evaluations are conducted on the standard UIEB~\cite{li2019underwater} and CSD~\cite{chen2021all} datasets, respectively. As shown in Table~\ref{tab:unseen_types}, DFC-IR consistently outperforms recent state-of-the-art methods across all evaluation metrics. For instance, DFC-IR yields notable PSNR improvements of 0.59 dB and 0.45 dB over the leading method MoCE-IR~\cite{zamfir2025complexity} on the UIEB and CSD datasets, respectively. These results indicate that DFC-IR generalizes beyond the degradation types observed during training. By decomposing degradation responses into band-wise spectral tokens and using them as frequency-aware restoration priors, DFC-IR handles unseen degradations through transferable local spectral response modeling, rather than relying on degradation-type labels or task-specific retraining. Furthermore, the visual comparisons in Fig.~\ref{fig_unseen_type} show that DFC-IR achieves more faithful correction of severe underwater color casts and more complete removal of dense snow particles than existing methods, further confirming its generalization ability under unseen degradation types.

\textbf{Unseen degradation levels.} To assess the generalization performance on unseen degradation levels, the evaluation is conducted under two settings. In the first setting, the five-degradation model is directly evaluated on the Rain100H \cite{yang2017deep} (heavy rain) dataset, with its deraining task trained exclusively on the Rain100L (light rain) dataset. The second setting examines the three-degradation model, originally trained with noise levels of $\sigma \in \{15, 25, 50\}$, on a more severe unseen noise condition of $\sigma = 75$. As presented in Table~\ref{tab:unseen_levels}, DFC-IR maintains superior performance when handling these unseen degradation levels. Specifically, DFC-IR achieves a PSNR gain of 0.45 dB over the leading method MoCE-IR \cite{zamfir2025complexity} on Rain100H, showing stronger transfer from light-rain training to heavy-rain testing. Under severe noise conditions, DFC-IR further surpasses VLU-Net \cite{zeng2025vision} by a substantial margin of 1.06 dB. These results demonstrate that DFC tokenization captures severity-aware degradation responses and enables DFC-IR to adapt band-wise restoration priors to unseen degradation intensities, rather than overfitting to the discrete degradation levels observed during training.

\begin{table}[!t]
\centering
\caption{Quantitative comparison of generalization performance on unseen degradation levels: heavy rain and severe noise.}
\label{tab:unseen_levels}
\setlength{\tabcolsep}{2.5pt} 
\resizebox{\linewidth}{!}{
\begin{tabular}{@{} l cccc cccc @{}}
\toprule
\multirow{2}{*}{Method} & 
\multicolumn{4}{c}{Rain100H} & 
\multicolumn{4}{c}{{BSD68}$_{\sigma=75}$}  \\
\cmidrule(lr){2-5} \cmidrule(lr){6-9} 
 & PSNR$\uparrow$ & SSIM$\uparrow$ & LPIPS$\downarrow$ & FID$\downarrow$ & PSNR$\uparrow$ & SSIM$\uparrow$ & LPIPS$\downarrow$ & FID$\downarrow$ \\
\midrule
Restormer \cite{zamir2022restormer} & 14.57 & 0.469 & 0.481 & 247.6 & 13.79 & 0.359 & 0.472 & 201.6 \\
PromptIR \cite{potlapalli2023promptir} & 14.69 & 0.478 & 0.471 & 243.4 & 18.46 & 0.401 & 0.406 & 170.1 \\
DFPIR \cite{tian2025degradation} & 15.56 & 0.494 & 0.403 & 197.5 & 18.67 & 0.416 & 0.392 & 164.1 \\
DA-RCOT \cite{tang2025degradation} & 15.84 & 0.521 & 0.381 & 170.4 & 20.63 & 0.474 & 0.358 & 140.6 \\
AdaIR \cite{cui2025adair} & 15.88 & 0.515 & 0.386 & 179.8 & 21.57 & 0.498 & 0.341 & 129.5 \\
VLU-Net \cite{zeng2025vision} & 15.97 & 0.524 & \underline{0.374} & \underline{164.2} & \underline{21.81} & \underline{0.506} & \underline{0.338} & \underline{120.6} \\
MoCE-IR \cite{zamfir2025complexity} & \underline{16.04} & \underline{0.531} & 0.397 & 184.6 & 20.18 & 0.467 & 0.376 & 156.1 \\
\midrule
DFC-IR (Ours) & \textbf{16.49} & \textbf{0.541} & \textbf{0.352} & \textbf{141.7} & \textbf{22.87} & \textbf{0.516} & \textbf{0.315} & \textbf{112.9} \\
\bottomrule
\end{tabular}
}
\end{table}

\subsection{Generalization to Real-world Scenarios}

\begin{figure*}[!t]
  \vspace{0cm}
  \centering
  \includegraphics[width=1\linewidth]{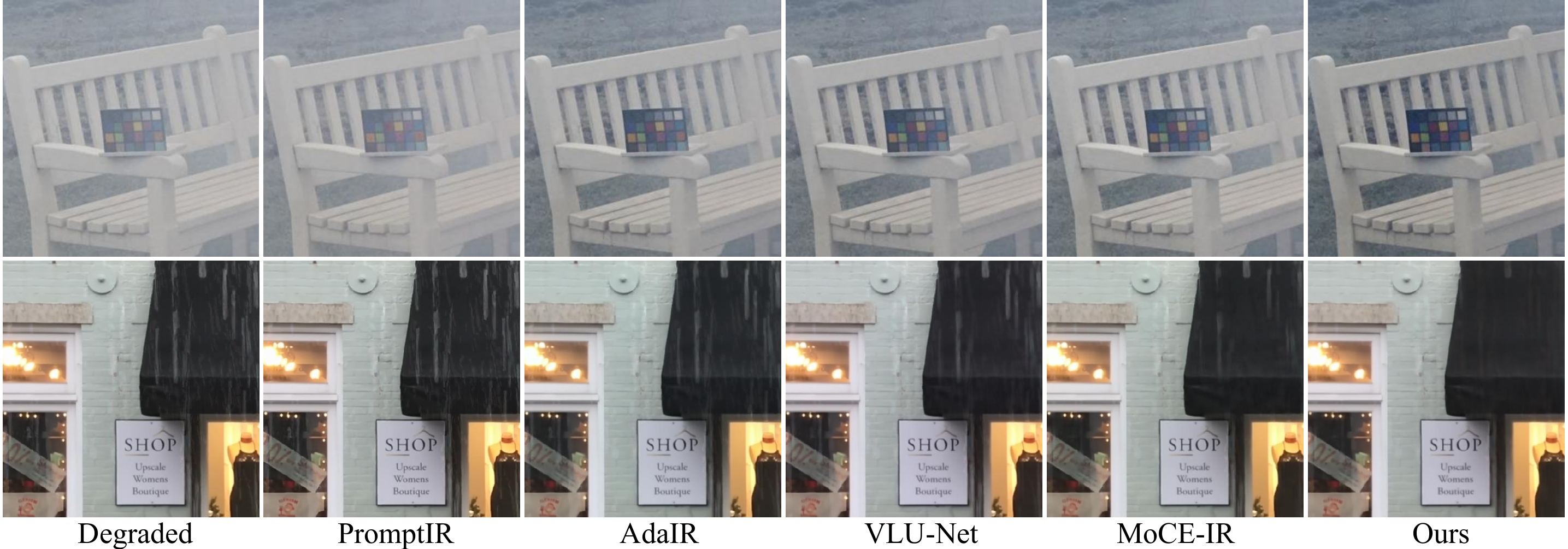}\\
  \vspace{0cm}
  \caption{Visual comparisons of generalization evaluation with five-degradation models on real-world NH-HAZE and SPANet. 
  }\label{fig_real_single}
  \vspace{0cm}
\end{figure*}

\textbf{Real-world individual degradations.} We first directly apply the five-degradation model to real-world datasets containing individual degradations. The evaluations are conducted on NH-HAZE \cite{ancuti2020nh} (dehazing), SPANet \cite{wang2019spatial} (deraining), and LOL-v2-real \cite{yang2021sparse} (low-light enhancement), utilizing PSNR, SSIM, LPIPS, and FID as quantitative metrics. As shown in Table~\ref{tab_real_single}, DFC-IR consistently outperforms existing approaches. Specifically, our model consistently outperforms the leading method MoCE-IR \cite{zamfir2025complexity} on NH-HAZE and SPANet, and notably surpasses the top-performing baseline DA-RCOT \cite{tang2025degradation} on LOL-v2-real with a remarkable PSNR margin of 0.95 dB and an FID reduction of 26.73. Furthermore, Fig.~\ref{fig_real_single} displays visual examples, revealing that while competing methods often leave residual degradations or introduce unnatural color shifts, our approach effectively restores faithful details and delivers improved perceptual quality.

\begin{table}[!t]
\centering
\caption{Quantitative comparison of five-degradation models on real-world individual degradations.}
\label{tab_real_single}
\setlength{\tabcolsep}{1.5pt} 
\resizebox{\linewidth}{!}{
\begin{tabular}{@{} l cccc cccc cccc @{}}
\toprule
\multirow{2}{*}{Method} & 
\multicolumn{4}{c}{NH-HAZE} & 
\multicolumn{4}{c}{SPANet} & 
\multicolumn{4}{c}{LOL-v2-real} \\
\cmidrule(lr){2-5} \cmidrule(lr){6-9} \cmidrule(lr){10-13}
 & PSNR$\uparrow$ & SSIM$\uparrow$ & LPIPS$\downarrow$ & FID$\downarrow$ & PSNR$\uparrow$ & SSIM$\uparrow$ & LPIPS$\downarrow$ & FID$\downarrow$ & PSNR$\uparrow$ & SSIM$\uparrow$ & LPIPS$\downarrow$ & FID$\downarrow$ \\
\midrule
Restormer \cite{zamir2022restormer} & 16.04 & 0.501 & 0.399 & 218.8 & 34.35 & 0.912 & 0.032 & 43.51 & 27.11 & 0.877 & 0.112 & 85.67 \\
PromptIR \cite{potlapalli2023promptir}  & 16.45 & 0.507 & 0.394 & 214.1 & 35.32 & 0.936 & 0.027 & 33.20 & 27.61 & 0.868 & 0.109 & 81.43 \\
DFPIR \cite{tian2025degradation} & 17.04 & 0.514 & 0.384 & 224.5 & 34.96 & 0.911 & 0.037 & 49.26 & 27.74 & 0.871 & 0.104 & 76.29 \\
DA-RCOT \cite{tang2025degradation} & 17.51 & 0.526 & 0.370 & 201.4 & 37.18 & 0.959 & \underline{0.019} & 20.77 & \underline{28.32} & \underline{0.905} & \underline{0.087} & \underline{65.68} \\
AdaIR \cite{cui2025adair} & 17.96 & 0.541 & 0.356 & 186.5 &36.99 & 0.951 & 0.022 & 23.02 & 28.09 & 0.899 & 0.095 & 70.11 \\
VLU-Net \cite{zeng2025vision} & 18.09 & 0.536 & 0.361 & 191.2 & 37.31 & 0.962 & \textbf{0.017} & \underline{19.96} & 27.25 & 0.881 & 0.110 & 88.49 \\
MoCE-IR \cite{zamfir2025complexity}   & \underline{18.17} & \underline{0.545} & \underline{0.350} & \underline{180.1} & \underline{37.54} & \textbf{0.971} & \textbf{0.017} & 24.18 & 28.18 & 0.902 & 0.106 & 84.57 \\
\midrule
DFC-IR (Ours) & \textbf{18.44} & \textbf{0.551} & \textbf{0.343} & \textbf{168.6} & \textbf{38.01} & \underline{0.969} & \textbf{0.017} & \textbf{19.12} & \textbf{29.27} & \textbf{0.929} & \textbf{0.062} & \textbf{38.95} \\
\bottomrule
\end{tabular}
}
\end{table}

\begin{table}[!t]
\centering
\caption{Quantitative evaluation of the five-degradation model on real-world composite degradations.}
\label{tab_real_comp}
\setlength{\tabcolsep}{4pt} 
\resizebox{\linewidth}{!}{
\begin{tabular}{@{} l ccc ccc @{}}
\toprule
\multirow{2}{*}{Method} & 
\multicolumn{3}{c}{Real haze and rain} & 
\multicolumn{3}{c}{Real blur and noise} \\
\cmidrule(lr){2-4} \cmidrule(lr){5-7} 
 & MANIQA$\uparrow$ & CLIPIQA$\uparrow$ & MUSIQ$\uparrow$ & MANIQA$\uparrow$ & CLIPIQA$\uparrow$ & MUSIQ$\uparrow$ \\
\midrule
Restormer \cite{zamir2022restormer} & 0.262 & 0.381 & 50.16 & 0.150 & 0.201 & 26.74 \\
PromptIR \cite{potlapalli2023promptir}  & 0.274 & 0.399 & 52.37 & 0.139 & 0.194 & 25.99 \\
DFPIR \cite{tian2025degradation}    & 0.268 & 0.395 & 54.27 & 0.147 & 0.198 & 26.91 \\
DA-RCOT \cite{tang2025degradation}     & 0.299 & 0.431 & 56.15 & 0.153 & 0.207 & 28.69 \\
AdaIR \cite{cui2025adair}     & 0.304 & 0.412 & 55.74 & 0.164 & 0.219 & 31.53 \\
VLU-Net \cite{zeng2025vision}     & 0.316 & \underline{0.457} & 57.74 & 0.168 & 0.225 & 32.27 \\
MoCE-IR \cite{zamfir2025complexity}   & \underline{0.324} & {0.448} & \underline{58.02} & \underline{0.177} & \underline{0.236} & \underline{33.38} \\
\midrule
DFC-IR (Ours) & \textbf{0.341} & \textbf{0.474} & \textbf{60.23} & \textbf{0.189} & \textbf{0.251} & \textbf{35.22} \\
\bottomrule
\end{tabular}
}
\end{table}

\begin{figure*}[!t]
  \vspace{0cm}
  \centering
  \includegraphics[width=1\linewidth]{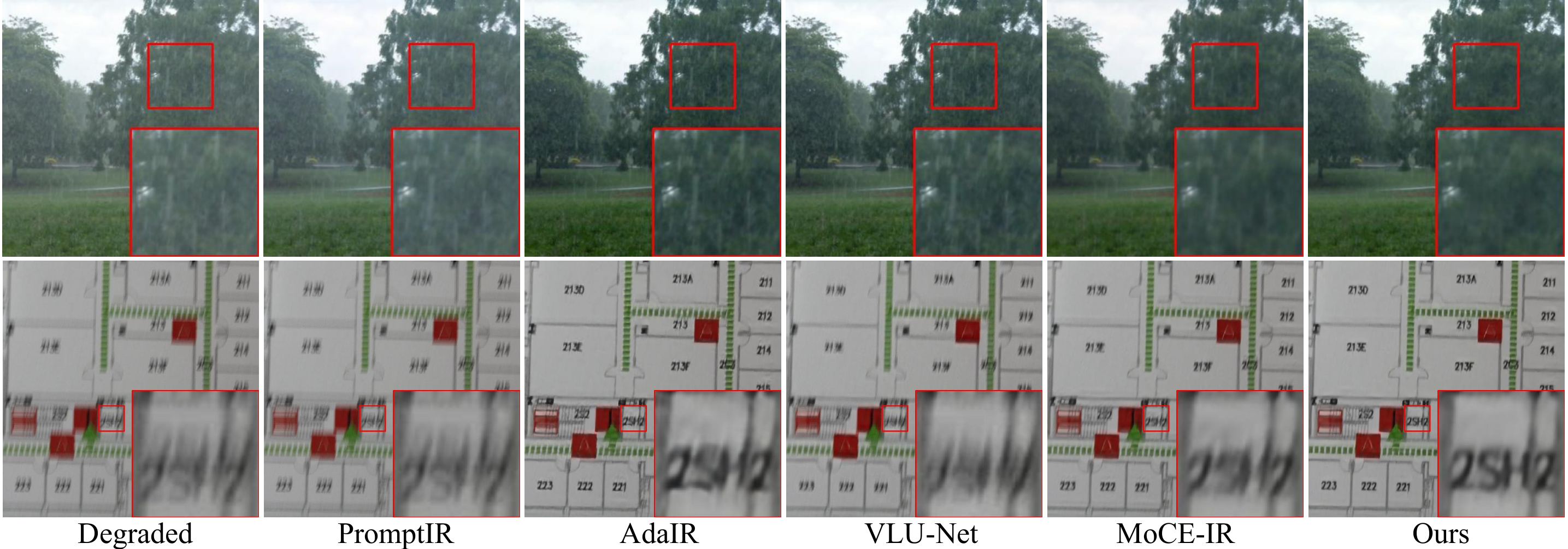}\\
  \vspace{0cm}
  \caption{Visual comparisons of generalization evaluation with five-degradation models on real-world composite degradations across haze/rain (row 1) and blur/noise (row 2).
  }\label{fig_real_comp}
  \vspace{0cm}
\end{figure*}

\textbf{Real-world composite degradations.} We subsequently extend the experiments to the real-world composite degradation setting. Specifically, the five-degradation model is evaluated on 45 real-world mixed-degradation images collected from Lai \cite{lai2016comparative} (blur and noise) and SPANet (rain and haze). For quantitative evaluation, we employ no-reference metrics including MANIQA, CLIPIQA, and MUSIQ. As presented in Table~\ref{tab_real_comp}, DFC-IR consistently attains the highest perceptual quality scores. For instance, our model surpasses the recent leading method MoCE-IR \cite{zamfir2025complexity} by yielding notable MUSIQ improvements of 2.21 on real haze and rain, and 1.84 on real blur and noise. Furthermore, visual comparisons in Fig.~\ref{fig_real_comp} illustrate that our method effectively clears rain and haze corruptions while retaining crucial scene details, and simultaneously resolves noise and blur without generating unnatural artifacts commonly seen in competing approaches. These results demonstrate that DFC tokenization provides decomposable band-wise priors for mixed real-world degradations, thereby significantly boosting the restoration capability under complex composite conditions.

\subsection{Ablation Studies}
We conduct ablation studies to validate the contribution of each key component in our proposed DFC-IR. Unless otherwise specified, the ablation experiments are performed in the all-in-one setting with three degradations.

\begin{table}[!t]
\centering
\caption{Ablation studies for the proposed components.}
\label{tab:architecture_modules}
\setlength{\tabcolsep}{6pt} 
\resizebox{\linewidth}{!}{ 
\begin{tabular}{@{} c cc cc cc @{}} 
\toprule
Method & FCMS & MMD & Params. & FLOPs & PSNR$\uparrow$ & SSIM$\uparrow$  \\
\midrule
(a) & \ding{55} & \ding{55} & 26.5M & 120.9G & 31.98 & 0.911  \\
(b) & \ding{51} & \ding{55} & 29.3M & 129.1G & 32.19 & 0.912  \\
(c) & \ding{55} & \ding{51} & 31.6M & 155.5G & 32.76 & 0.916  \\
\midrule
Ours & \ding{51} & \ding{51} & 33.1M & 163.7G & \textbf{{33.00}} & \textbf{{0.919}}  \\
\bottomrule
\end{tabular}
}
\end{table}

\textbf{Impact of individual architecture modules}. Table~\ref{tab:architecture_modules} summarizes the performance benefits of individual architectural contributions. As observed in variant (b), the proposed FCMS module brings a gain of 0.21 dB PSNR over the baseline model (a) by adaptively converting the scale-specific DFC into band-wise spectral tokens through DFC-guided frequency partitioning. Similarly, variant (c) demonstrates that employing the MMD module alone yields a 0.78 dB PSNR improvement, highlighting the effectiveness of the multi-band decoding structure in performing band-specific feature modulation and adaptive aggregation for degradation-aware restoration. Ultimately, the full model integrates FCMS for DFC-guided band-wise tokenization and MMD for DFC-token-conditioned feature modulation and aggregation, achieving a substantial 1.02 dB overall gain. This validates the complementary roles of explicit DFC tokenization and token-conditioned band-specific feature processing in improving degradation-aware restoration.

\begin{table*}[!t]
\centering
\caption{Ablation study of the multi-scale (MS) supervision and the frequency-domain loss $\mathcal{L}_{f}$.}
\label{tab:Aba_loss}
\setlength{\tabcolsep}{1.2em} 
\begin{tabular}{cccc cc cc cc cc}
\toprule
\multirow{2}{*}{Variant} & \multirow{2}{*}{$\mathcal{L}_s$} & \multirow{2}{*}{MS} & \multirow{2}{*}{$\mathcal{L}_{f}$} & \multicolumn{2}{c}{SOTS} & \multicolumn{2}{c}{Rain100L} & \multicolumn{2}{c}{BSD68$_{15-50}$} & \multicolumn{2}{c}{Average} \\
\cmidrule(lr){5-6} \cmidrule(lr){7-8} \cmidrule(lr){9-10} \cmidrule(lr){11-12}
 & & & & PSNR$\uparrow$ & SSIM$\uparrow$ & PSNR$\uparrow$ & SSIM$\uparrow$ & PSNR$\uparrow$ & SSIM$\uparrow$ & PSNR$\uparrow$ & SSIM$\uparrow$ \\
\midrule
(a) & \checkmark & $\times$ & $\times$ & 30.41 & 0.971 & 37.26 & 0.972 & 31.15 & 0.872 & 32.22 & 0.912 \\
(b) & \checkmark & $\times$ & \checkmark & 30.85 & 0.977 & 37.51 & 0.975 & 31.18 & 0.873 & 32.38 & 0.914 \\
(c) & \checkmark & \checkmark & $\times$ & 31.82 & 0.980 & 38.34 & 0.982 & 31.29 & 0.874 & 32.81 & 0.917 \\
(d) Ours & \checkmark & \checkmark & \checkmark & \textbf{32.28} & \textbf{0.982} & \textbf{38.74} & \textbf{0.985} & \textbf{31.33} & \textbf{0.876} & \textbf{33.00} & \textbf{0.919} \\
\bottomrule
\end{tabular}
\end{table*}

\begin{table}[!t]
\centering
\caption{Ablation for different numbers of band-wise DFC tokens.}
\label{tab:frequency_bands} 
\setlength{\tabcolsep}{3pt} 
\resizebox{\linewidth}{!}{ 
\begin{tabular}{@{} cccccccccc @{}} 
\toprule
DFC   & 
\multirow{2}{*}{Params.} & 
\multicolumn{2}{c}{SOTS} & 
\multicolumn{2}{c}{Rain100L} & 
\multicolumn{2}{c}{{BSD68}$_{15-50}$} & 
\multicolumn{2}{c}{Average}  \\
\cmidrule(lr){3-4} \cmidrule(lr){5-6} \cmidrule(lr){7-8} \cmidrule(lr){9-10}
Tokens& & PSNR$\uparrow$ & SSIM$\uparrow$ & PSNR$\uparrow$ & SSIM$\uparrow$ & PSNR$\uparrow$ & SSIM$\uparrow$ & PSNR$\uparrow$ & SSIM$\uparrow$ \\
\midrule
2 & 30.6M & 31.21 & 0.977 & 38.34 & 0.980 & 31.05 & 0.869 & 32.54 & 0.913 \\
3 & 31.8M & 31.69 & 0.978 & 38.47 & 0.981 & 31.18 & 0.873 & 32.74 & 0.916 \\
4 (ours) & 33.1M & \textbf{32.28} & \textbf{0.982} & \textbf{38.74} & \textbf{0.985} & \textbf{31.33} & \textbf{0.876} & \textbf{33.00} & \textbf{0.919} \\
5 & 34.4M & 32.25 & 0.981 & 38.71 & 0.984 & 31.32 & 0.874 & 32.98 & 0.917 \\
\bottomrule
\end{tabular}
}
\end{table}

\begin{table}[!t]
\centering
\caption{Ablation studies between uniform and DFC-guided tokenization strategies on unseen degradations.}
\label{tab:Aba_token}
\setlength{\tabcolsep}{2.5pt} 
\resizebox{\linewidth}{!}{
\begin{tabular}{@{} l cccc cccc @{}}
\toprule
\multirow{2}{*}{Tokenization} & 
\multicolumn{4}{c}{UIE (UIEB)} & 
\multicolumn{4}{c}{Desnowing (CSD)}  \\
\cmidrule(lr){2-5} \cmidrule(lr){6-9} 
 & PSNR$\uparrow$ & SSIM$\uparrow$ & LPIPS$\downarrow$ & FID$\downarrow$ & PSNR$\uparrow$ & SSIM$\uparrow$ & LPIPS$\downarrow$ & FID$\downarrow$ \\
\midrule
Uniform & 18.76 & 0.809 & 0.251 & 30.99 & 20.51 & 0.774 & 0.328 & 90.41 \\
DFC-guided (Ours) & \textbf{19.24} & \textbf{0.817} & \textbf{0.231} & \textbf{27.76} & \textbf{21.05} & \textbf{0.802} & \textbf{0.291} & \textbf{70.68}\\
\bottomrule
\end{tabular}
}
\end{table}

\begin{figure}[!t]
  \vspace{0cm}
  \centering
  \includegraphics[width=1\linewidth]{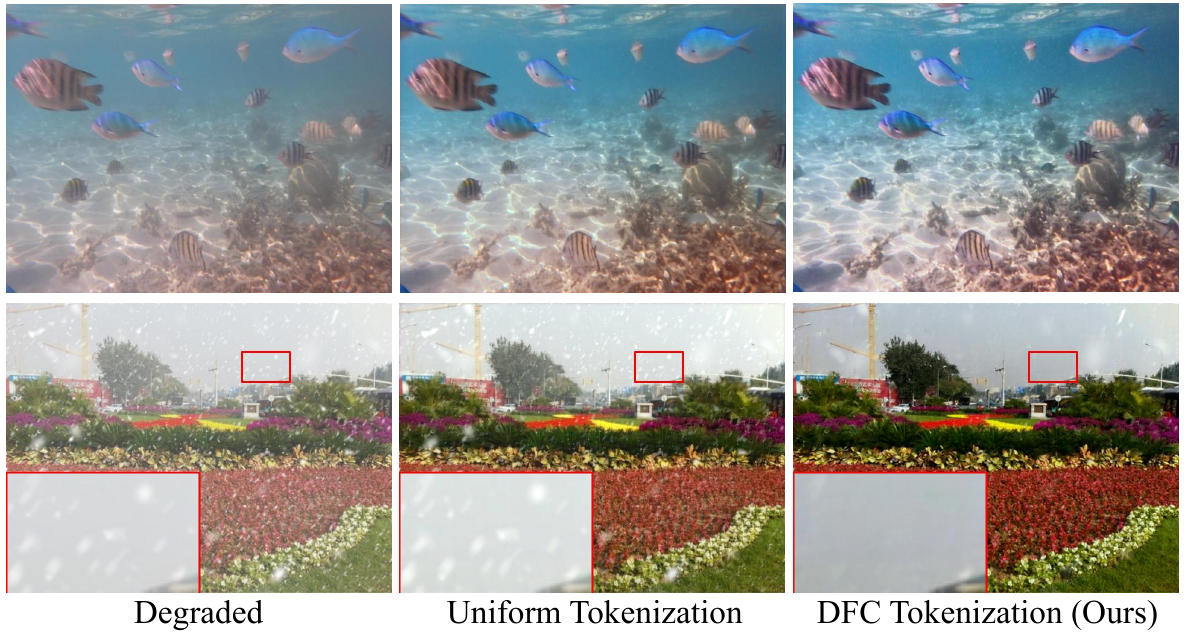}\\
  \vspace{0cm}
  \caption{Visual results of uniform tokenization and our DFC-guided tokenization on unseen degradations. Compared with the uniform baseline, our DFC-guided tokenization reconstructs details with enhanced fidelity.
  }\label{fig:Aba_token}
  \vspace{0cm}
\end{figure}

\textbf{Effects of loss functions.} We conduct an ablation study to validate the effectiveness of the multi-scale (MS) supervision and the frequency-domain loss $\mathcal{L}_{f}$. As presented in Table \ref{tab:Aba_loss}, adding $\mathcal{L}_{f}$ in variant (b) yields a marginal average PSNR improvement of only 0.16 dB over the baseline (a). Without explicit MS constraints, the under-constrained intermediate restoration $\hat{\bm{x}}^s$ introduces errors into scale-specific DFC estimation, resulting in suboptimal tokenization. Conversely, applying MS supervision with only the spatial $\mathcal{L}_s$ loss in (c) stabilizes the intermediate outputs for more reliable DFC estimation, yielding a significant PSNR gain of 0.59 dB over the baseline. Finally, our full model (d) integrates both MS supervision and $\mathcal{L}_{f}$, surpassing the baseline by 0.78 dB. These results indicate that MS supervision plays a critical role in stabilizing DFC estimation, while $\mathcal{L}_{f}$ further improves the spectral consistency of DFC-token-conditioned restoration outputs.

\textbf{Number of band-wise DFC tokens}. As shown in Table~\ref{tab:frequency_bands}, partitioning the DFC into 4 band-wise tokens yields optimal performance, achieving the highest average PSNR of 33.00 dB. This configuration strikes the best balance between spectral granularity and model efficiency. Specifically, it provides sufficient resolution to distinguish local DFC responses across different frequency regions, such as high-frequency responses dominated by noise and middle-to-high-frequency responses caused by rain streaks. Meanwhile, it avoids redundant tokens and unnecessary band-wise modulation branches caused by over-partitioning. In comparison, coarse partitioning with 2 or 3 tokens yields suboptimal restoration results because overly broad frequency segments merge heterogeneous degradation responses and reduce token discriminability. Conversely, extending the setting to 5 tokens increases the parameter count to 34.4M but leads to a slight performance drop, indicating that over-partitioning increases parameter redundancy and complicates token aggregation.

\textbf{Effect of tokenization strategy.} We evaluate the proposed DFC-guided tokenization strategy by comparing it against a uniform tokenization baseline. The ablation is conducted using the five-degradation model on two unseen degradation tasks: underwater image enhancement (UIEB) and desnowing (CSD). As presented in Table \ref{tab:Aba_token}, our DFC-guided strategy consistently outperforms the uniform baseline on both datasets. For instance, it yields substantial PSNR improvements of 0.48 dB on UIEB and 0.54 dB on CSD. Furthermore, visual comparisons in Fig. \ref{fig:Aba_token} corroborate these results. The uniform strategy struggles with unknown degradations, leaving severe underwater color shifts (row 1) and distinct residual snow particles (row 2). In contrast, our DFC-guided strategy effectively removes these degradations and restores faithful details. This demonstrates that DFC-guided tokenization better aligns token boundaries with sample-specific DFC responses than rigid uniform division, producing more effective band-wise DFC tokens for unseen degradations. These tokens further enable token-conditioned band-specific feature modulation and adaptive aggregation in MMD, thereby achieving robust generalization to unseen degradations.

\textbf{Reliability of DFC estimation.} As described in Sec.~\ref{sec:4.1}, the same DFC estimation pipeline is used in both training and inference, where the DFC at each scale is computed from the intermediate restoration $\hat{\bm{x}}^s$ and the degraded input. To further assess its reliability, we construct an oracle variant by replacing $\hat{\bm{x}}^s$ with the corresponding GT image only for DFC computation. As presented in Table~\ref{tab:estimated_vs_gt}, the performance gap is negligible, with only a 0.01 dB PSNR gain when using GT-derived DFC. This is attributed to the multi-scale DFC-token-conditioned supervision, which directly constrains intermediate restorations and thereby stabilizes scale-specific DFC estimation.

\begin{table}[!t]
\centering
\caption{Performance comparison of DFC estimation using the intermediate restoration $\hat{\bm{x}}^s$ versus the GT.}
\label{tab:estimated_vs_gt}
\renewcommand{\arraystretch}{1}
\setlength{\tabcolsep}{2pt} 
\resizebox{\linewidth}{!}{
\begin{tabular}{@{} l cccccccc @{}} 
\toprule
\multirow{2}{*}{Method} & 
\multicolumn{2}{c}{SOTS} & 
\multicolumn{2}{c}{Rain100L} & 
\multicolumn{2}{c}{BSD68$_{15-50}$} & 
\multicolumn{2}{c}{Average} \\
\cmidrule(lr){2-3} \cmidrule(lr){4-5} \cmidrule(lr){6-7} \cmidrule(lr){8-9}
 & PSNR$\uparrow$ & SSIM$\uparrow$ & PSNR$\uparrow$ & SSIM$\uparrow$ & PSNR$\uparrow$ & SSIM$\uparrow$ & PSNR$\uparrow$ & SSIM$\uparrow$ \\
\midrule
w/ GT DFC & 32.28 & 0.982 & 38.76 & 0.986 & 31.34 & 0.876 & 33.01 & 0.919 \\
w/ Est. DFC (ours) & 32.28 & 0.982 & 38.74 & 0.985 & 31.33 & 0.876 & 33.00 & 0.919 \\
\bottomrule
\end{tabular}
}
\end{table}

\subsection{Efficiency Analysis}
We evaluate the computational efficiency of DFC-IR in terms of parameters, FLOPs, GPU memory, and inference latency. As shown in Table~\ref{tab:Efficiency}, DFC-IR achieves the best restoration quality while maintaining competitive efficiency. Compared with DFPIR, DFC-IR improves PSNR by 0.39 dB, while reducing GPU memory and inference latency by 41.8\% and 12.3\%, respectively, at the cost of 6.4\% more parameters and 8.3\% more FLOPs. Compared with VLU-Net, DFC-IR achieves a 0.92 dB PSNR gain with 6.5\% fewer parameters, 2.6\% fewer FLOPs, 45.8\% less GPU memory, and 4.0\% lower latency. This efficiency is primarily attributed to the DFC-token-based architectural design: FCMS uses a lightweight network to adaptively generate band-wise DFC tokens, while MMD leverages token-conditioned cross-attention and adaptive aggregation for band-specific restoration, avoiding heavy sub-networks.

\begin{table}[!t]
\centering
\small
\caption{Computational efficiency under the five-degradation setting. Evaluated on one NVIDIA A800 GPU with an input size of $256 \times 256$.}
\label{tab:Efficiency}

\setlength{\extrarowheight}{3pt} 
\setlength{\tabcolsep}{3.5pt} 

\resizebox{\linewidth}{!}{
\begin{tabular}{ c  cc  cc  cc } 
\hline 
Method & Params. & FLOPs & GPU Mem. & Latency & PSNR$\uparrow$ & SSIM$\uparrow$ \\
\hline 
Restormer \cite{zamir2022restormer} & 26.1M & 141.0G & 667.5M & 85.77ms & 27.60 & 0.881 \\
PromptIR \cite{potlapalli2023promptir} & 35.6M & 158.1G & 711.5M & 81.52ms & 29.15 & 0.904 \\
VLU-Net \cite{zeng2025vision} & 35.4M & 168.1G & 762.8M & 127.7ms & 30.11 & 0.905 \\
AdaIR \cite{cui2025adair} & 28.8M & 147.2G & 678.3M & 92.25ms & 30.20 & 0.910 \\
DA-RCOT \cite{tang2025degradation} & 52.3M & 333.1G & 800.7M & 220.2ms & 30.40 & 0.911 \\
MoCE-IR \cite{zamfir2025complexity} & 25.4M & 86.25G & 314.9M & 67.21ms & 30.58 & 0.919 \\
DFPIR \cite{tian2025degradation} & 31.1M & 151.1G & 709.7M & 139.8ms & 30.64 & 0.913 \\
\hline
DFC-IR (Ours) & 33.1M & 163.7G & 413.1M & 122.6ms & \textbf{31.03} & \textbf{0.922} \\
\hline
\end{tabular}
}
\end{table}

\section{Conclusion}
In this paper, we introduced DFC, an explicit, quantifiable, and interpretable spectral representation for characterizing degradation responses in the frequency domain. To translate DFC into restoration guidance, we further developed DFC-IR, which decomposes scale-specific DFCs into band-wise spectral tokens and leverages these tokens as degradation-aware priors for coarse-to-fine restoration. Extensive experiments on standard all-in-one benchmarks, composite degradations, unseen degradations, and real-world scenarios demonstrate that the proposed framework achieves state-of-the-art restoration performance and strong generalization under complex degradation conditions. These results highlight the effectiveness of DFC as an explicit degradation representation and suggest its potential as a practical foundation for all-in-one image restoration.

\bibliography{references}
\vfill
\bibliographystyle{IEEEtran}

\end{document}